\documentclass[10pt,journal,letterpaper,compsoc]{IEEEtran}

\ifCLASSOPTIONcompsoc
\else
\fi
\usepackage{longtable}
\usepackage{amsfonts}
\usepackage{algorithmic}
\usepackage{algorithm}
\usepackage[pdftex]{graphicx}
 \DeclareGraphicsExtensions{.pdf,.jpeg,.png,.eps}

\begin{document}
%

\title{Extended Dynamic Programming and \\ Fast Multidimensional Search Algorithm \\ for Energy Minization in Stereo and Motion }


\author{Mikhail~G.~Mozerov~
\IEEEcompsocitemizethanks{\IEEEcompsocthanksitem M. Mozerov is with the Computer Vision Center 
of Department Informatics, Universitat Autònoma de Barcelona, Barcelona,
Spain, 08193.\protect\\
E-mail: mozerov@cvc.uab.es}
\thanks{}}


\markboth{ARXIV publication}%
{Shell \MakeLowercase{\textit{et al.}}: Extended Dynamic Programming and Fast Multidimensional Search Algorithm for Energy Minization in Stereo and Motion}


\IEEEcompsoctitleabstractindextext{%
\begin{abstract}
This paper presents a novel extended dynamic programming approach for energy minimization (EDP) to solve the correspondence problem for stereo and motion. A significant speedup is achieved using a recursive minimum search strategy (RMS). The mentioned speedup is particularly important if the disparity space is 2D as well as 3D. The proposed RMS can also be applied in the well-known dynamic programming (DP) approach for stereo and motion. In this case, the general 2D problem of the global discrete energy minimization is reduced to several mutually independent sub-problems of the one-dimensional minimization. The EDP method is used  when the approximation of the general 2D discrete energy minimization problem is considered. Then the RMS algorithm is an essential part of the EDP method. Using the EDP algorithm we obtain a lower energy bound than the graph cuts (GC) expansion technique on stereo and motion problems. The proposed calculation scheme possesses natural parallelism and can be realized on  graphics processing unit (GPU) platforms, and can be potentially restricted further by the number of scanlines in the image plane. Furthermore, the RMS and EDP methods can be used in any optimization problem where the objective function meets specific conditions in the smoothness term.
\end{abstract}

}

\maketitle

\IEEEdisplaynotcompsoctitleabstractindextext
\IEEEpeerreviewmaketitle

\section{Introduction}
\IEEEPARstart{S}{tereo} and motion matching are used in many applications and remains one of the most challenging open problems in computer vision \cite{bib:Scharstein,bib:Brown}.  
The most successful algorithms apply a global correspondence strategy based on an energy-minimization framework \cite{bib:Szeliski,bib:Szeliski2}.

Historically, DP was one of the first attempts to solve the stereo matching problem by minimizing the global energy \cite{bib:Gimelfarb, bib:Kanade}. It was called scanline optimization
when the global 2D problem was split into a many one dimensional sub-problems \cite{bib:Zitnik,bib:Veksler2}. Practically, the same DP approach was applied for the motion correspondence problem when a a multidimensional disparity space was considered \cite{bib:Sun1, bib:Gong2}. 
In spite of state-of-the-art results in dense disparity map reconstruction \cite{bib:Hirschmuller, bib:Wang, bib:Zhang,bib:Mattoccia, bib:Salmen} the level of global energy minimization of the DP optimization was not satisfactory due to the loss of several essential prior dependencies in the smoothness term of the discrete energy function, e.g. the vertical dependencies in stereo matching in  scanline optimization.

To obtain an exact solution of the energy minimization problem, the GC method for linear prior dependencies was proposed by Roy and Cox in \cite{bib:Roy}. This approach was extended for convex prior dependencies by Ishikawa in \cite{bib:Ishikawa}. The computational complexity of both algorithms in  case of an exact solution is rather high, especially the Ishikawa approach. Therefore, many  GC techniques that aim to obtain an approximate solution have been proposed \cite{bib:Boykov, bib:Kolmogorov, bib:Kolmogorov2,bib:Boykov2,bib:Komodakis, bib:Veksler}. 
 
Among the GC techniques, we give special consideration to the GC expansion method \cite{bib:Boykov} for two main reasons. The first is that the expansion algorithm still is very popular in computer vision community because it is relatively fast and  obtains excellent approximate solutions to the energy minimization problem (only a few algorithms were reported to achieve lower energy level \cite{ bib:Veksler,bib:Kolmogorov3, bib:Kumar, bib:Kumar2}; but the computational complexity in these cases is higher than in \cite{bib:Boykov}). The second is that  Middlebury provides an open C++ code for the expansion algorithm which we use to make comparisons with our EDP approximation. In general,  GC techniques cannot be implemented for parallel calculation schemes and there is no theoretical guarantee that the parallelization will be faster for every problem instance \cite{bib:Strandmark}.
In contrast, belief propagation (BP) \cite{bib:Kolmogorov3, bib:LBP1, bib:LBP2,bib:Yang} and DP approaches possess a natural parallelism and can be realized on the GPU platform, as done in \cite{bib:Pock}, for example.

The general stereo and motion correspondence problem assumes a multidimensional disparity space (usually 2D or 3D domain). In this case, the computational complexity of straightforward search of DP algorithms is extremely high \cite{bib:Sun1, bib:Sun2, bib:Gong2, bib:Wang}, and a fast algorithm is highly desirable. For this purpose, we have designed a RMS algorithm that reduces the computational complexity of a straightforward search. 

The proposed algorithm results in a considerable complexity reduction - that is, $O\left( {N{Q^2}} \right) \to O\left( {NWQ} \right) \to O\left( {NQ} \right)$, for an image with $N$ pixels, and a search domain cardinality $Q $. A constant $W$ is the cardinality of a constrained search subdomain of the RMS algorithm: $W << Q $. We have to note that the solution of the fast RMS algorithm is the same as the solution of the straightforward search. 
To show that the DP approach still remains competitive even for energy minimization problem, we propose in this paper an improved extension of DP.  The so-called EDP algorithm aims to obtain an approximate solution and the method inherits ideas that were proposed in \cite {bib:Mozerov, bib:Mozerov2}.  However, using this new modification of the DP approach we obtain a lower energy than the expansion algorithm \cite{bib:Boykov} on stereo and motion problems. We have to note that the EDP is a kind of DP, which ignores the uniqueness constraint like other approaches based on energy minimization do (GC, BP, etc.), and unlike DP with ordering does.

The paper is organized as follows: In Section 2 the problem is formulated. In Section 3 we describe the fast RMS algorithm. The EDP algorithm is described in Section 4. Computer experiments of the proposed techniques are discussed in Section 5. Section 6 summarizes our conclusions.  

\section{Problem Definition}

\begin{figure}[!t]
\centering
\includegraphics[width=3.25in]{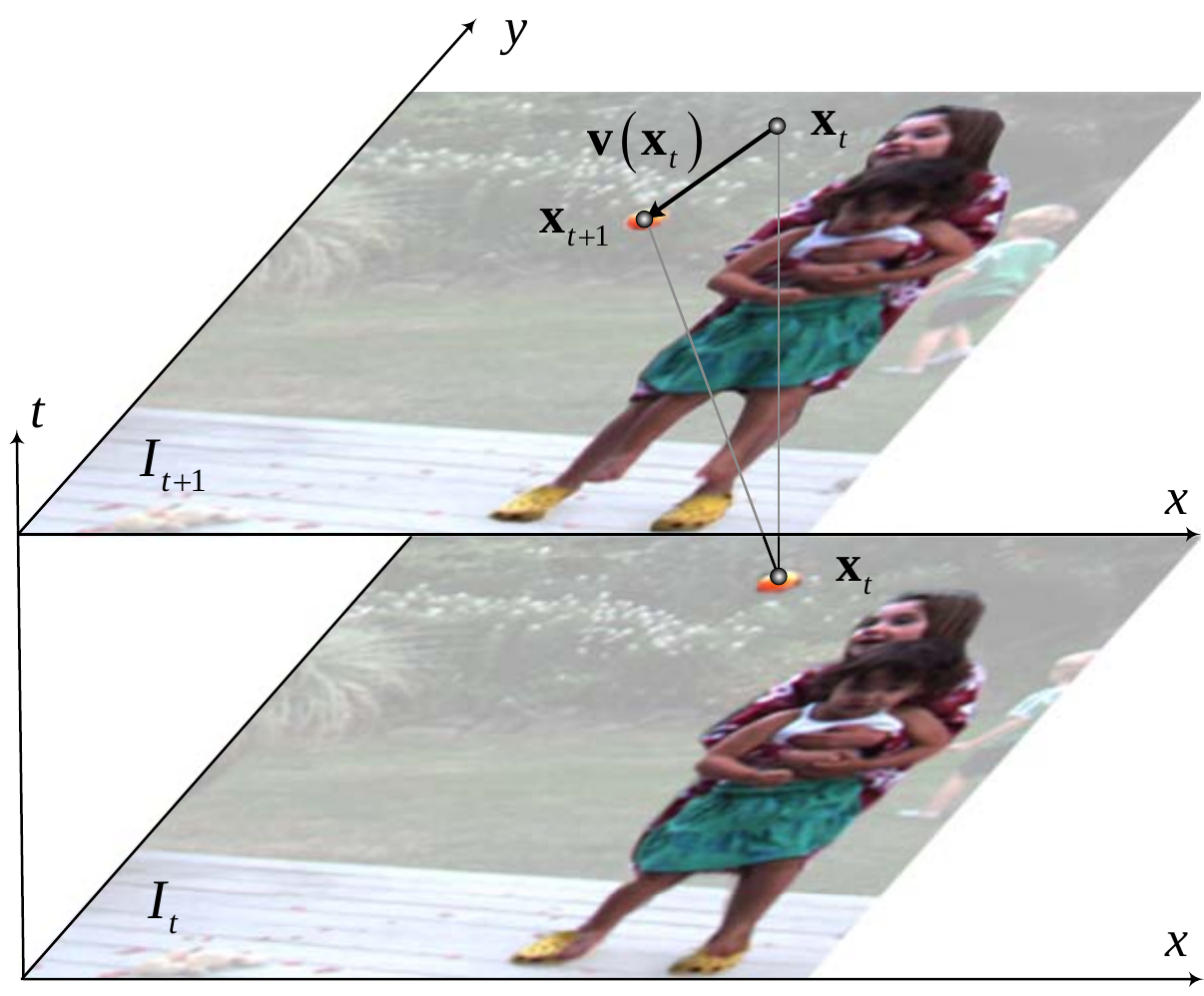}
\caption{Scheme of correspondence matching between two images.}
\label{fg:MEscheme}
\end{figure}

The general stereo and motion matching approach aims to find correspondence between pixels of  images $I_t \left( {\mathbf{x}} \right)$ and $I_{t + 1} \left( {\mathbf{x}} \right)$ as it is shown in Fig.~\ref {fg:MEscheme}, where ${\mathbf{x}}$ is a coordinate of a pixel in the image plane, $t$ is an index of the considered image in the sequence. A vector ${\mathbf{v}}\left( {{\mathbf{x}}_t } \right)$
in Fig.~\ref {fg:MEscheme} denotes the disparity of two corresponding pixels ${\mathbf{x}}_t $ and ${\mathbf{x}}_{t+1} $. In general, the image point coordinate $ \mathbf {x}$ is a multidimensional vector, it is 2D in the case of stereo and dynamic images, 3D in the case of a tomography image. The disparity vector $ \mathbf {v}$ has usually the same dimensionality as the domain of the image except in certain special cases: for instance, if stereo matching is considered the disparity vector domain becomes one dimensional due to the additional epipolar constraints. Simply expressed,  if the stereo or motion matching problem is considered a dense disparity map $\mathbf {v}( \mathbf {x})$ has to be obtained. The question is how to solve such a problem in the most appropriate way.
\begin{figure}[!t]
\centering
\includegraphics[width=3.25in]{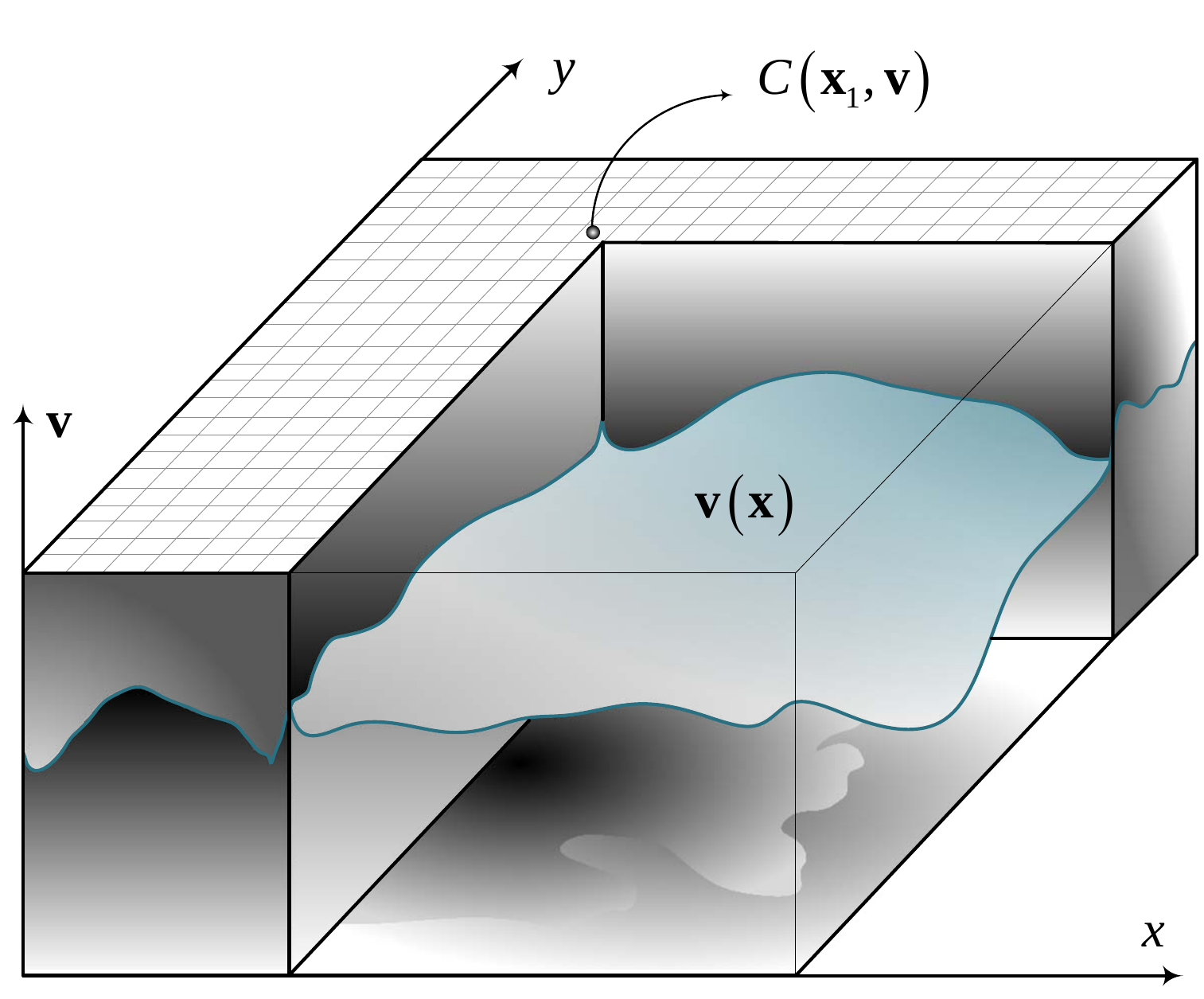}
\caption{The DSI as a collection of correspondence costs $C\left( {{\mathbf{x}},{\mathbf{v}}} \right)$ and a desired function of disparity value $\mathbf {v}( \mathbf {x})$.}
\label{fg:DSIscheme}
\end{figure}
The global energy minimization approach intends to find a desired disparity function $\mathbf {v}( \mathbf {x})$, which minimizes the energy function $E(\mathbf {v}(\mathbf {x}))$ in the disparity space image (DSI) $C(\mathbf {x},\mathbf {v})$, see Fig.~\ref {fg:DSIscheme}. The DSI represents a collection of correspondence cost. For example, if two compared pixels $\left( {{\mathbf{x}}_1 } \right)_t $ and $\left( {{\mathbf{x}}_1 + {\mathbf{v}}_1} \right)_{t + 1} $ have the same luminance value (which means that these pixels are a potential match) the cost value $C\left( {{\mathbf{x}}_1 ,{\mathbf{v}}_1} \right)$ might be minimal. Vice versa, if the luminance values differ, the related cost value increases. The global energy usually contains two terms, the data term and the smoothness term
\begin{equation}
E\left( {{\bf{v}}({\bf{x}})} \right) = \sum\limits_{{\bf{x}} \in \Omega } {C\left( {{\bf{x}},{\bf{v}}({\bf{x}})} \right)} {\rm{  +  }}\sum\limits_{{\bf{x}} \in \Omega } {G\left( {{\bf{v}}({\bf{x}})} \right)} {\rm{,}} 
\label{eq:GE_objective_function}
\end{equation}
where $G$ is a smoothness function and  ${\mathbf{\Omega }}$ is the domain of the vector ${\mathbf{x}} = \left\{ {{x_1},{x_2},..,{x_D}} \right\}$ or $ \left\{ {{x},{y},..,{z}} \right\}$ with the cardinality $\left| {\mathbf{\Omega }} \right| = N$. The domain of the vector ${\mathbf{v}} = \left\{ {{v_1},{v_2},..,{v_R}} \right\}$ is denoted as $\mathbf{V}$ with the cardinality $\left| {\mathbf{V}} \right| = Q$. The image domain that we have to work with is a discrete vector space, and it is reasonable to consider vector components $x_d$ and $v_r$ as proportional to the natural numbers (e.g. $\alpha \left( {{v_r} + v_r^0} \right),\beta \left( {{x_d} + x_d^0} \right) \in \mathbb{N}$). Since $\alpha$, $\beta$, $v_r^0$ and $x_d^0$ can be any arbitrary real constants, our discretization model allows subpixel accuracy. If the discretization step in the disparity domain is less than in the image domain, the accuracy of the disparity estimation is obviously at subpixel level. Thus, a substitution of such variables for the natural scale is trivial, and further in this paper we will assume that the natural numeration of the indexes in the multidimensional vector grid coincide with the values of the discrete vectors $ \mathbf {x}$ and $ \mathbf {v}$. Consequently, the conditions of the mentioned vector domain is
\begin{equation}
\begin{array}{l}
 {v_r} \in \left\{ {1,2,..,v_r^{\max }} \right\},v_r^{\max } \in \mathbb{N}, \hfill \\
 {x_d} \in \left\{ {1,2,..,x_d^{\max }} \right\},x_d^{\max } \in \mathbb{N}. \hfill \\ 
\end{array}
\label{eq:v_r_x_d}\end{equation}
The cost values $C(\mathbf {v}(\mathbf {x}))$ that form a DSI might be computed as follows
\begin{equation}
C\left( {{\bf{x}},{\bf{v}}} \right){\rm{ = }}{\left| {{I_{{\rm{t  +  1}}}}({\bf{x}} + {\bf{v}}) - {I_{\rm{t}}}({\bf{x}})} \right|^{{\rm{1}}{\rm{,2}}}} \wedge {C_{\max }}{\rm{,}}
\label{eq:Cost_value}\end{equation}
where $I_{t+1}$ and $I_t$ are luminance values of two neighboring images in stereo or dynamic sequences, $ \wedge $ is a sign of the pairwise $\mathbf{min}$ value operation, $C_{\max }$ is a cost truncation constant. We also use $ \bigwedge $ as a sign of $\mathbf{min}$ value operation defined on a set of variables. The cost truncation constant $C_{\max }$ is introduced in our paper for some practical reasons: this parameter is referred in the experimental section.  If $C_{\max } = \infty $ then the comparison metric (\ref{eq:Cost_value}) is the most popular $L_1$ or $L_2$ norm. Nonetheless, there are many other cost metrics (see e.g. \cite{bib:Scharstein,bib:Brown}) that could be used for the DSI formation.

The function $G$ in the smoothness term of (\ref{eq:GE_objective_function}) is given by
\begin{equation}
G\left( {{\bf{v}}({\bf{x}})} \right){\rm{  =  }}\lambda \sum\limits_{d \in D} {\sum\limits_{r \in R} {f\left( {\left| {{{\rm{v}}_r}\left( {{x_d} + 1} \right) - {{\rm{v}}_r}\left( {{x_d}} \right)} \right|} \right)} } ,
\label{eq:Smooth_value}\end{equation}
where $\lambda $ is a constant used to penalize motion vector discontinuities, $R$ and $D$ are the dimensionalities of motion vector and image spaces respectively. Later on in this paper a shortcut 
\begin{equation}
\begin{array}{l}
 f\left( {\left| {{\bf{v}}\left( {{x_d} + 1} \right) - {\bf{v}}\left( {{x_d}} \right)} \right|} \right) =  \\ 
 \sum\limits_{r \in R} {f\left( {\left| {{{\rm{v}}_r}\left( {{x_d} + 1} \right) - {{\rm{v}}_r}\left( {{x_d}} \right)} \right|} \right)}  \\ 
 \end{array}
\label{eq:Smooth_value}\end{equation}
is used for the distance measure. A positive definite increasing function $f$ is usually proportional to the gradient of the motion vector or its squared value
\begin{equation}
G\left( {{\mathbf{v}}\left( {\mathbf{x}} \right)} \right) = \lambda \sum\limits_{d \in D} {\left| {{\mathbf{v}}\left( {x_d  + 1} \right) - {\mathbf{v}}\left( {x_d } \right)} \right|} ^{1,2}, \label{eq:Smooth_value_l}\end{equation}
To prevent over-penalizing discontinuity a more flexible smoothness function is used
\begin{equation}
\bar G({\mathbf{v}}({\mathbf{x}})) = \lambda \sum\limits_{d \in D} {f\left( {|{\mathbf{v}}(x_d + 1) - {\mathbf{v}}(x_d )|} \right) \wedge \lambda f(g),} 
\label{eq:Smooth_value_t}\end{equation}
where $g$ is a truncation threshold and $\lambda \left( {\mathbf{x}} \right)$  is locally adaptive in such away that $\lambda \left( {\mathbf{x}} \right) = 2\lambda $ if a value of a local gradient of an image $I_t \left({\mathbf{x}} \right)$ is less than 10 and $\lambda \left( {\mathbf{x}} \right) = 2\lambda $ otherwise. The meaning of this threshold becomes clearer if we consider disparity map estimation in stereo for the most popular truncated linear prior, e.g. the case of $f\left( {\left| a \right|} \right) = \left| a \right|$ in (\ref {eq:Smooth_value_t}). Indeed, the smoothness rate is proportional to the disparity difference and matters only inside an object segment, where this difference is supposed to be relatively small. In contrast, there is no reason to penalize a solution more for being 20 instead of 5, for example, if it is known that both values belong to the same class of discontinuity: say foreground object - background surface. We suppose that the positive definite threshold $g$ belongs to the domain as ${\left| {v_r } \right|}$, in other words is a natural number. 

\begin{figure}[!t]
\centering
\includegraphics[width=3.30in]{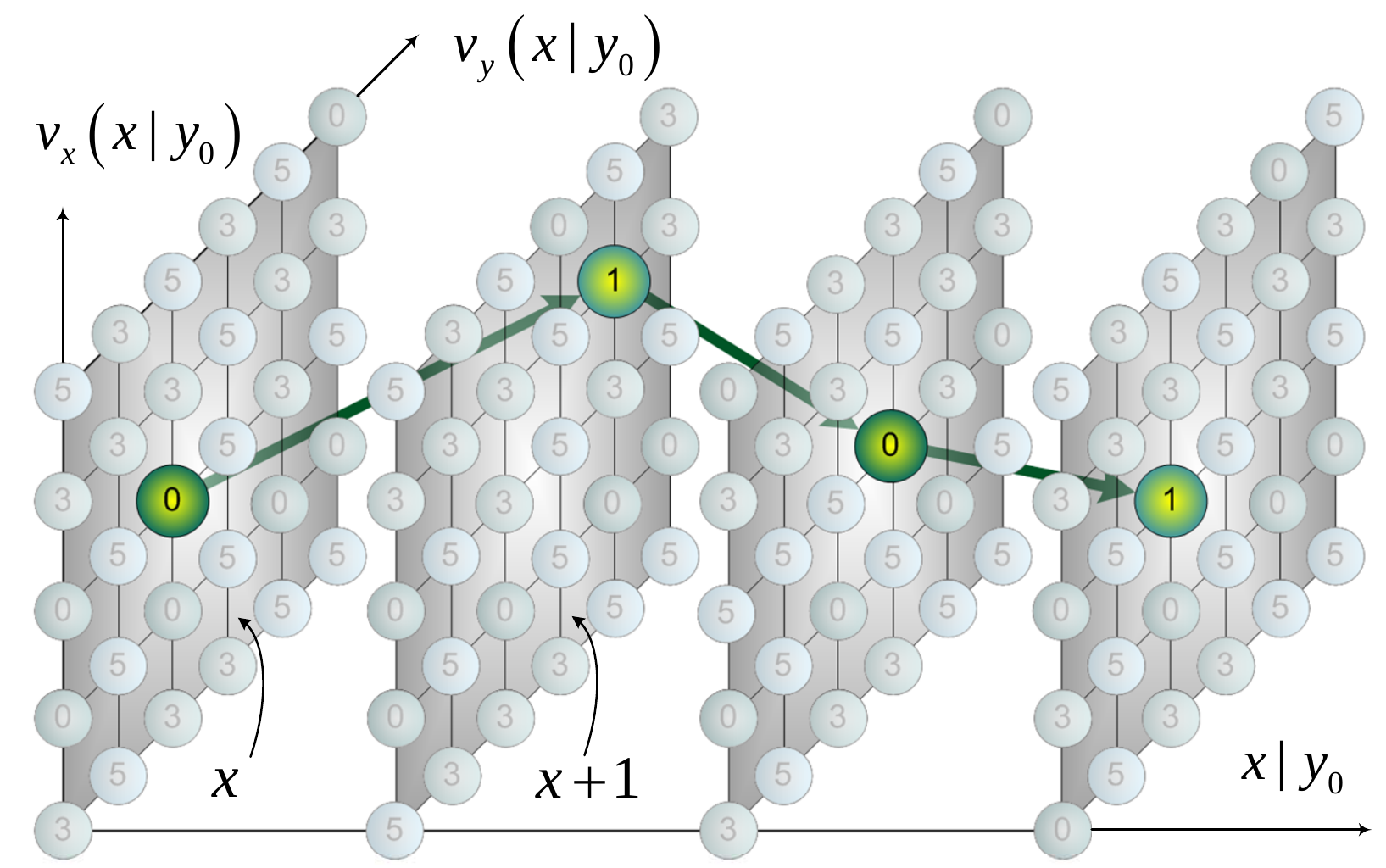}

\centering (a)

\centering
\includegraphics[width=3.30in]{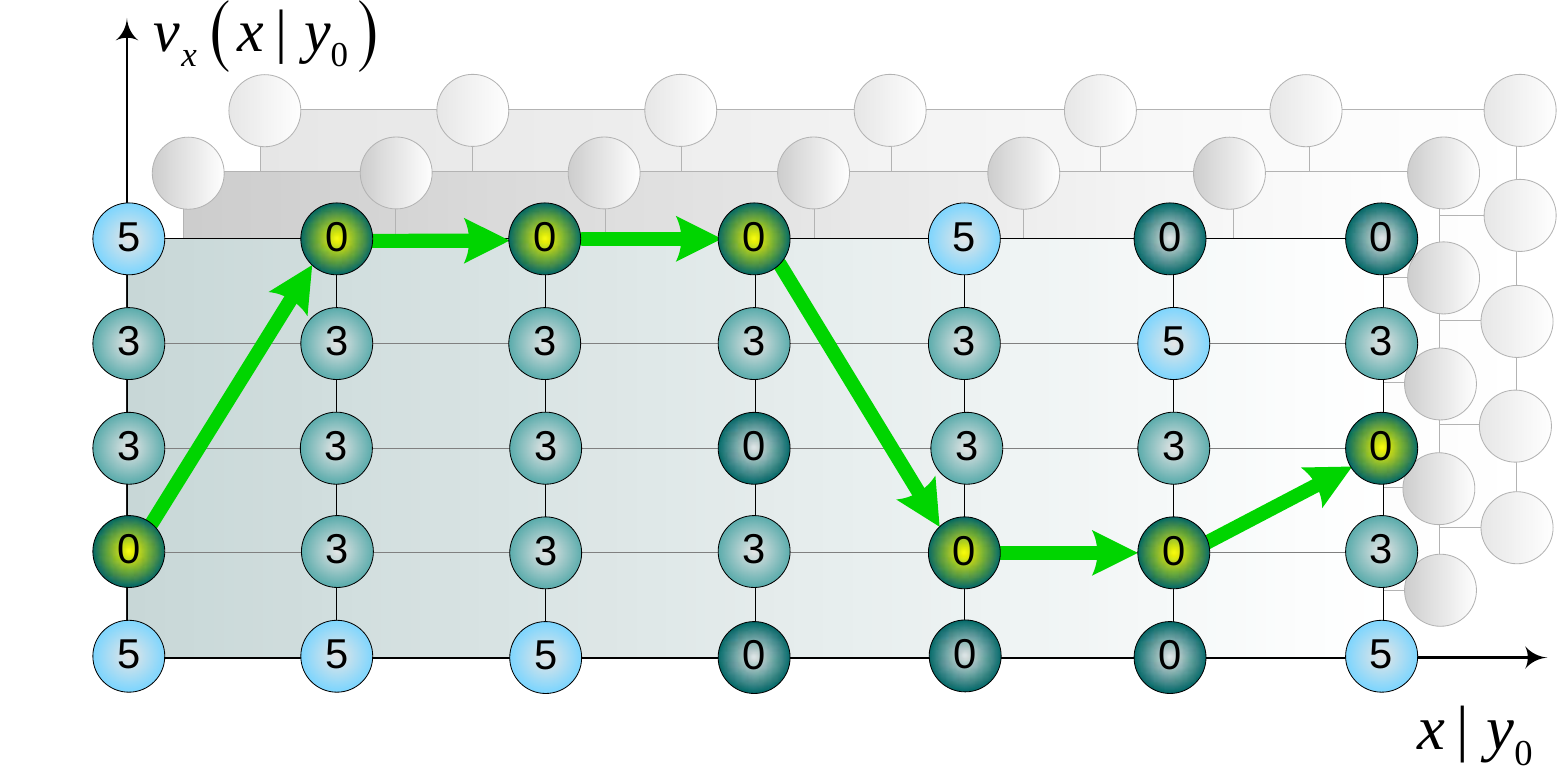}

\centering (b)

\caption{The optimal path through the DSI trellis, (a) for 2D motion, (b) for stereo.}
\label{fg:DP_1D}
\end{figure}

The general 2D problem of the global energy minimization can be reduced to several mutually independent sub-problems of the one-dimensional minimization if several additional constraints of the smoothness term are considered, e.g. the vertical prior dependencies are equal to zero. Then, the solution is given by
\begin{equation}
{{\mathbf{\bar v}}}\left( {x|y_0 } \right) = \mathop {\arg \min }\limits_{\mathbf{v}} \left\{ \hspace{-0.05 in} \begin{array}{l}
 \sum\limits_{x \in \Omega _x } {C\left( {x,y_0 ,{\mathbf{v}}\left( {x|y_0 } \right)} \right) + } \hfill \\
  \lambda \hspace{-0.06 in}\sum\limits_{x \in \Omega _x } {\hspace{-0.06 in} \left( {f\left( {\left| {{\mathbf{u}}\left( {x|y_0 } \right)} \right|} \right) \wedge f\left( g \right)} \right)} \hfill \\ 
\end{array} \hspace{-0.05 in} \right\},
\label{eq:One_d}\end{equation}
where $y_0$ is the fixed index of the considered sub-problem and $f\left( {\left| {{\mathbf{u}}(x|y_0 )} \right|} \right) = f\left( {\left| {{\mathbf{v}}(x + 1|y_0 ) - {\mathbf{v}}(x|y_0 )} \right|} \right)$. Each discrete function ${\mathbf{v}}(x|y_0)$ coincides with the optimal (minimal energy) path through the 3D trellis (Fig.~\ref {fg:DP_1D}(a)) for motion, which is a slice of the initial 4D DSI. The optimal path for stereo is illustrated in Fig.~\ref {fg:DP_1D}(b). Hereafter, the fixed index $y_0$ of (\ref {eq:One_d}) is omitted for notational simplicity.  

\begin{figure}[!t]
\centering
\includegraphics[width=3.30in]{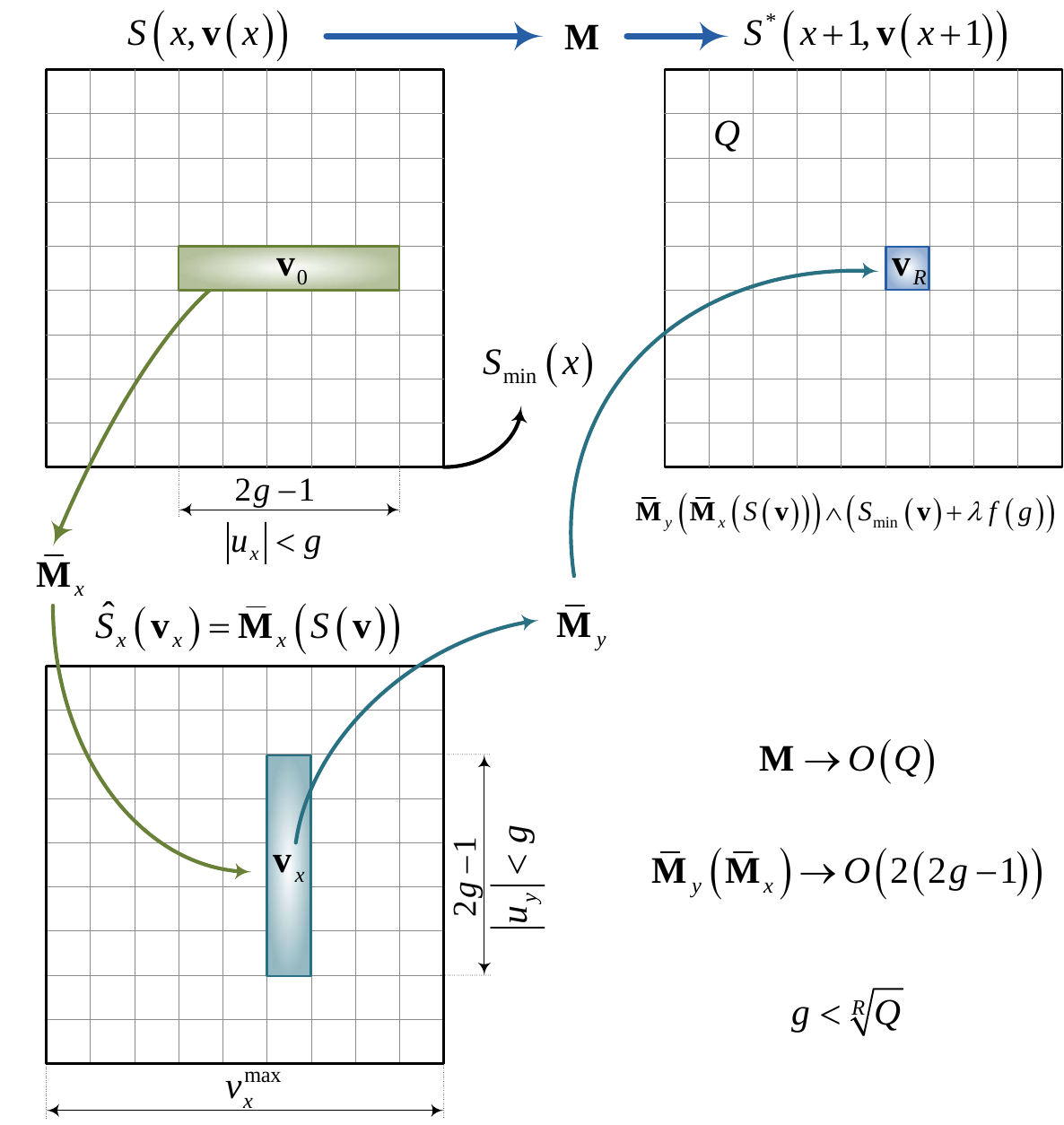}
\caption{The straightforward minimum search versus the RMS algorithm.}
\label{fg:Cut_wnd}
\end{figure}

Such problems can be solved by means of the DP algorithm. The method consists of step-by-step control and optimization.  
Let $S\left( {x,{\mathbf{v}}\left( x \right)} \right)$ be the optimal path cost to vertex $\left( {x,{\mathbf{v}}\left( x \right)} \right)$, see Fig.~\ref {fg:DP_1D}. Then the optimal path cost to vertex $\left( {x+1,{\mathbf{v}}\left( x+1 \right)} \right)$ is given by the recurrence relation
\begin{equation}
\begin{array}{l}
 S\left( {x + 1,{\mathbf{v}}\left( {x + 1} \right)} \right) = C\left( {x + 1,{\mathbf{v}}\left( {x + 1} \right)} \right) + \hfill \\
 \hspace{-0.06 in} \underbrace {\mathop \bigwedge \limits_{{\mathbf{v}}\left( x \right)}\hspace{-0.06 in} S\left( {x,{\mathbf{v}}\left( x \right)} \right) + \lambda f\left( {\left| {{\mathbf{v}}\left( {x + 1} \right) - {\mathbf{v}}\left( x \right)} \right|} \right) \wedge \lambda f\left(g\right)}_{S^* \left( {x + 1,{\mathbf{v}}\left( {x + 1} \right)} \right)}. \hfill \\ 
\end{array}
\label{eq:base}\end{equation}
Equation (\ref {eq:base}) is an essential part of any DP based technique. It is obvious that the calculation of the sum $ S^* \left( {x + 1,{\mathbf{v}}\left( {x + 1} \right)} \right)$ in (\ref {eq:base}) is the most time consuming operation. Therefore, let ${\mathbf{M}}\left( {S\left( {x,{\mathbf{v}}\left( x \right)} \right),\bar G\left( {x,{\mathbf{v}}\left( x \right)} \right)} \right) = {\mathbf{M}}\left( {S\left( {x,{\mathbf{v}}\left( x \right)} \right)} \right)$ be the operator that transforms the full array of optimal cost $S\left( {x,{\mathbf{v}}\left( x \right)} \right)$ at the step $x$ (Fig.~\ref {fg:DP_1D} and Fig.~\ref {fg:Cut_wnd}) to the additional sum $ S^* \left( {x + 1,{\mathbf{v}}\left( {x + 1} \right)} \right)$ at the step $x+1$  
\begin{equation}
{S^*}\left( {x + 1,{\bf{v}}\left( {x + 1} \right)} \right) = {\bf{M}}\left( {S\left( {x,{\bf{v}}\left( x \right)} \right)} \right)
\label{eq:M_strf}
\end{equation}
Then, the computational complexity of the operator $\mathbf{M}$ that is used in step-by-step optimization (\ref {eq:base}) via (\ref {eq:M_strf}) is proportional to $O\left( {Q } \right)$ per each vertex $\left( {\mathbf{x}},{\mathbf{v}} \right)$ in the DSI space, see illustration in Fig.~\ref {fg:Cut_wnd}, where $Q$ is the number of labels in the multidimensional disparity space for the energy minimization problem.

Therefore, the main idea of the RMS approach is to make the DP step-by-step operator $\mathbf{M}$ significantly   faster, and this idea will be explained in the next section.  

\section{Recursive Search Algorithm}

This section is subdivided into two subsections. The first subsection explains the general case of the RMS algorithm, and the second subsection introduces the RMS technique for the truncated linear prior, because in this case a further speedup can be obtained.
\subsection{General RMS Algorithm}
By the distributive and associative laws of the minimum-addition semi-ring over reals (or integers) one can get from (\ref {eq:base}) 
\begin{equation}
\begin{array}{*{20}{c}}

   {{S^*}\left( {x + 1,{\bf{v}}\left( {x + 1} \right)} \right) = }  \\
   {\underbrace {\mathop {MIN}\limits_{{\bf{v}}\left( x \right):f\left( {\left| {{\bf{u}}\left( x \right)} \right|} \right) < f\left( g \right)} \left( {\begin{array}{*{20}{c}}
   {S\left( {x,{\bf{v}}\left( x \right)} \right) + }  \\
   {\lambda f\left( {\left| {{\bf{u}}\left( x \right)} \right|} \right){\rm{  }}}  \\
\end{array}} \right) \wedge }_{\hat S\left( {x + 1,{\bf{v}}\left( {x + 1} \right)} \right)}{\rm{ }}}  \\
   {\underbrace {\mathop {MIN}\limits_{{\bf{v}}\left( x \right)} \left( {S\left( {x,{\bf{v}}\left( x \right)} \right) + \lambda f\left( g \right)} \right)}_{{S_{\min }}\left( x \right) + \lambda f\left( g \right)},{\rm{ }}}  \\
\end{array}
\label{eq:derivation1}\end{equation}
in which we used a shortcut ${\bf{u}}\left( x \right) = {\bf{v}}\left( {x + 1} \right) - {\bf{v}}\left( x \right)$.

The term $\hat S\left( {x + 1,{\mathbf{v}}\left( {x + 1} \right)} \right)$ says that one has a natural limit on the minimum search range. The term of  $S_{\min } \left( x \right) = \mathop \bigwedge \limits_{{\mathbf{v}}\left( x \right)} \left( {S\left( {x,{\mathbf{v}}\left( x \right)} \right)} \right)$ is easy to obtain as a by-product of the previous phase of the forward pass of the DP algorithm. In fact, assuming $ f\left( g \right) > 0$, the $f\left( g \right)$ in $f\left( {\left| {{\mathbf{u}}\left( {x } \right) } \right|} \right) < f\left( g \right)$ in (\ref {eq:derivation1}) can be replaced with $ f\left( {\left| {{\mathbf{u}}\left( {x } \right) } \right|} \right) < af\left( g \right)$ where $a \ge 1$  is a real constant. This merely corresponds to enlarging the search range. The value of  $S^* $ will not change if we choose some $a > 1$ due to clipping by the value of $S_{\min } \left( x \right) + \lambda f\left( g \right)$.   This helps us decompose the term $\hat S$ recursively, per dimension: There is always some finite $a \ge 1$  so that
\begin{equation}
\begin{array}{l}
 \hat S\left( {x + 1,{\bf{v}}\left( {x + 1} \right)} \right) =  \\ 
 \mathop {MIN}\limits_{f\left( {\left| {{\bf{u}}\left( x \right)} \right|} \right) < af\left( g \right)} \left\{ {S\left( {x,{\bf{v}}\left( x \right)} \right) + \lambda \sum\limits_{i = 1}^R {f\left( {\left| {{u_i}\left( x \right)} \right|} \right)} } \right\} =  \\ 
 \mathop {MIN}\limits_{f\left( {\left| {{u_R}} \right|} \right) < f\left( g \right)} ...\mathop {MIN}\limits_{f\left( {\left| {{u_1}} \right|} \right) < f\left( g \right)} \left( \begin{array}{l}
 S\left( {x,{\bf{v}}\left( x \right)} \right) + \lambda f\left( {\left| {{u_1}} \right|} \right) +  \\ 
 \lambda f\left( {\left| {{u_2}} \right|} \right) + ... + \lambda f\left( {\left| {{u_R}} \right|} \right) \\ 
 \end{array} \right) =  \\ 
 \mathop {MIN}\limits_{\left| {{u_R}} \right| < g} {\rm{ }}..\mathop {MIN}\limits_{\left| {{u_2}} \right| < g} \mathop {MIN}\limits_{\left| {{u_1}} \right| < g} \left( {S + \lambda f\left( {\left| {{u_1}} \right|} \right)} \right) + \lambda f\left( {\left| {{u_2}} \right|} \right) \\ 
 \end{array}  
\label{eq:derivation2}\end{equation}
in which we used a shortcut $u_i = v_i \left( {x + 1} \right) - v_i \left( x \right)$ and $f\left( {\left| {u_i } \right|} \right) < f\left( g \right)$, thus the $f$ is a positively definite  increasing function by our definition.

Therefore, the per dimensional decomposition of the straightforward minimum search in (\ref {eq:derivation2}) is the key property the RMS algorithm, which gives significant speedup of the sum $S^* \left( {x + 1,{\mathbf{v}}\left( {x + 1} \right)} \right)$ calculation in the step-by-step optimization in (\ref {eq:base}). Now, let 
\begin{equation}
\begin{array}{l}
 \hat S_r \left( {{\mathbf{v}}_r } \right) = {\mathbf{\bar M}}_r \left( {\hat S_{r - 1} \left( {{\mathbf{v}}_{r - 1} } \right)} \right) = \hfill \\
\mathop \bigwedge \limits_{\left| {u_r } \right| < g} \left( {\hat S_{r - 1} \left( {{\mathbf{v}}_{r - 1} } \right) + \lambda f\left( g \right)} \right), \hfill \\ 
\end{array}  
\label{eq:M_decmp1}\end{equation}
then from (\ref {eq:derivation2})
\begin{equation}
\begin{array}{l}
 S^* \left( {x + 1,{\mathbf{v}}\left( {x + 1} \right)} \right) = {\mathbf{M}}\left( {S\left( {x,{\mathbf{v}}\left( x \right)} \right)} \right) = \hfill \\ \hfill \\
 {\mathbf{\bar M}}_R \left( \hspace{-0.01 in}{.\hspace{-0.015 in}.\hspace{-0.015 in}.{\mathbf{\bar M}}_2 \hspace{-0.025 in}\left( {{\mathbf{\bar M}}_1 \hspace{-0.025 in}\left( {S \hspace{-0.02 in}\left( {x,{\mathbf{v}}\left( x \right)} \hspace{-0.02 in}\right)} \hspace{-0.02 in}\right)}\hspace{-0.02 in} \right)} \hspace{-0.02 in}\right)\hspace{-0.02 in} \wedge \hspace{-0.02 in}\left( {S_{\min }\hspace{-0.025 in} \left( x \right)\hspace{-0.017 in} + \hspace{-0.017 in}\lambda f\hspace{-0.025 in}\left( g \right)} \hspace{-0.02 in}\right)\hspace{-0.015 in}, \hfill \\ 
\end{array} 
\label{eq:M_decmp}\end{equation}
where the ${\mathbf{v}}_r$ is an intermediate domain
\begin{equation}
{\mathbf{v}}_r = \left\{ {v_R \left( x \right),...v_{r + 1} \left( x \right),v_r \left( {x + 1} \right),...,v_1 \left( {x + 1} \right)} \right\},
\label{eq:M_decmp2}\end{equation}
introduced to make sense of the variable $u_r$ in (\ref {eq:derivation2}), (\ref {eq:M_decmp1}) and is used for calculation of the intermediate function $\hat S_r \left( {{\mathbf{v}}_r } \right)$. Then obviously $\hat S_R \left( {{\mathbf{v}}_R } \right) = \hat S\left( {x + 1,{\mathbf{v}}\left( {x + 1} \right)} \right)$ and $\hat S_0 \left( {{\mathbf{v}}_0 } \right) = S\left( {x,{\mathbf{v}}\left( x \right)} \right)$.

The application of the fast calculation scheme derived in (\ref {eq:derivation1}) and (\ref {eq:derivation2}) for the 2D case is illustrated in Fig.~\ref {fg:Cut_wnd}, in which the sum $\hat S_x $ denotes an intermediate results of the application of the operator ${\mathbf{\bar M}}_x $.
The computational complexity of each one-dimensional operator ${\mathbf{\bar M}}_r$ is $2g-1$ operations per each vertex $\left( {\mathbf{x}},{\mathbf{v}} \right)$ in the DSI space. The application of  all $R$ operators gives $R(2g-1)$ per each vertex $\left( {\mathbf{x}},{\mathbf{v}} \right)$ instead of $Q$. Note that the maximum value of the threshold $g$ in general is $g <<\sqrt[R]{Q}$, and this is a significant speedup especially in the multidimensional case shown in the experimental section of the paper. 
If the smoothness term is a truncated linear prior and $g>2$, further speedup can be achieved as it is shown in the next subsection.

\subsection{RMS Algorithm for Truncated Linear Prior}

The value of  $S^* $ will not change if we replace a domain $v_i :\left| {u_i } \right| < g$
 in (\ref {eq:derivation2}) by $v_i :\left| {u_i } \right| \le v_r^{\max }< \infty $ due to clipping by the value of $S_{\min } \left( x \right) + \lambda f\left( g \right)$. Taking the linear prior $f\left( {\left| {u_r } \right|} \right) = \left| {u_r } \right|$ we can replace ${\mathbf{\bar M}}_r $ in (\ref {eq:M_decmp}) by an operator ${\mathbf{L}}_r $, which is defined as  
\begin{equation}
\begin{array}{l}
{\mathbf{L}}_r \hspace{-0.03 in}\left(\hspace{-0.01 in} {\hat S_{r - 1}\hspace{-0.01 in} \left(\hspace{-0.02 in} {\mathbf{v}_{r-1}} \hspace{-0.01 in}\right)} \hspace{-0.01 in}\right) = \mathop \bigwedge \limits_{{\left| {u_r } \right|} \le v_r^{\max } } \hspace{-0.03 in}\left(\hspace{-0.02 in} {\hat S_{r - 1} \hspace{-0.02 in}\left(\hspace{-0.02 in} {\mathbf{v}_{r-1}} \hspace{-0.02 in}\right)\hspace{-0.02 in} + \hspace{-0.025 in}\lambda {\left| {u_r } \right|}} \hspace{-0.02 in}\right)\hspace{-0.02 in}. \hfill \\ 
\end{array}  
\label{eq:L_dec}\end{equation}
\begin{figure}[!t]
\centering
\includegraphics[width=3.39in]{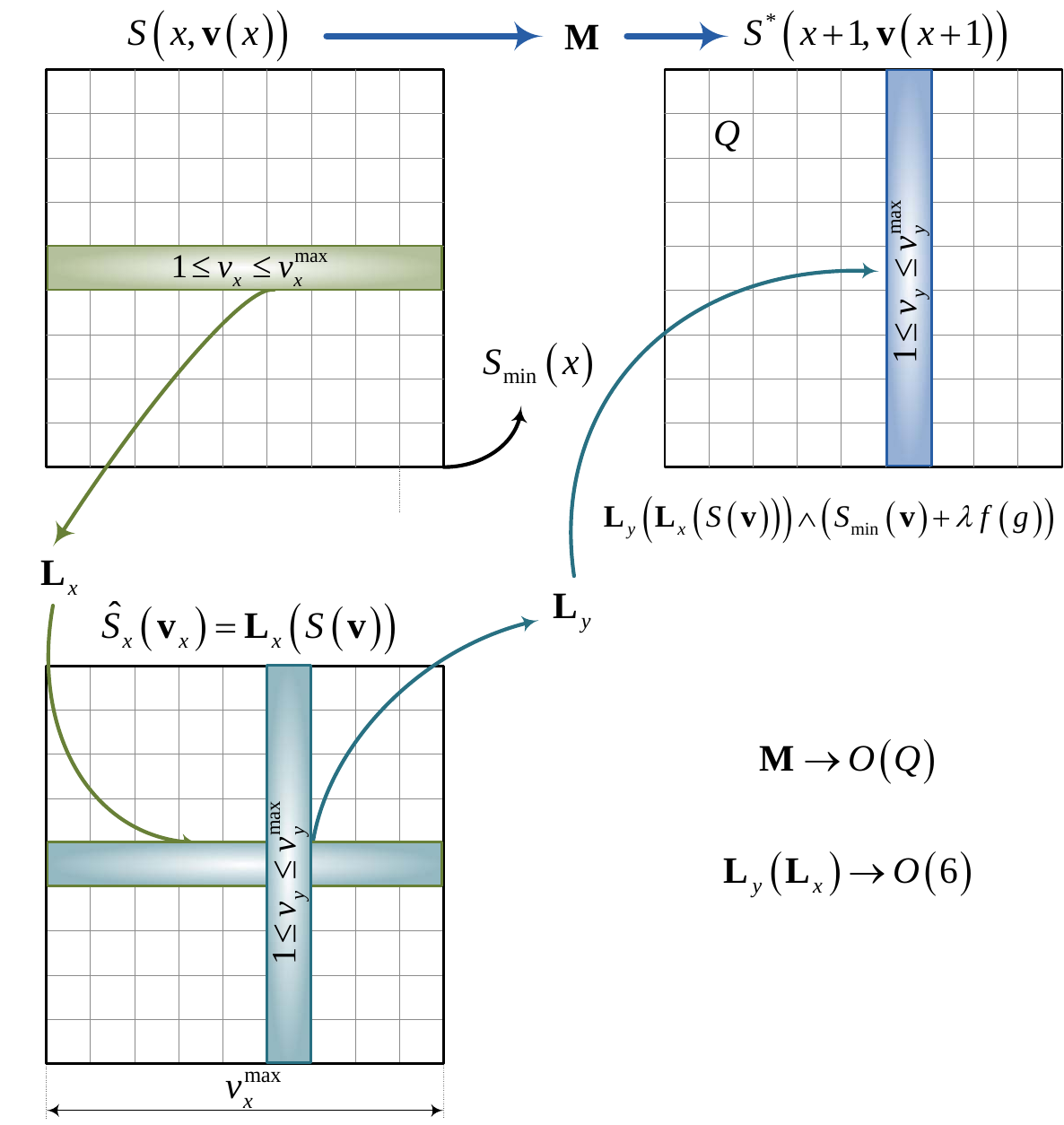}
\caption{The straightforward minimum search versus the RMS algorithm for the truncated linear prior.}
\label{fg:Cut_wnd_l}
\end{figure}
Let us rewrite (\ref {eq:L_dec}) by taking $v' = v_r \left( {x + 1} \right)$, $v = v_r \left( x \right)$, $u = v' - v$ and omitting some unimportant components ${\mathbf{v}}_r /v'$ and ${\mathbf{v}}_{r - 1} /v$ of the vector domains ${\mathbf{v}}_r $ and ${\mathbf{v}}_{r - 1} $, by using the distributive and associative laws of the minimum-addition semi-ring and property of the absolute value operation 
\begin{equation}
\begin{array}{l}
 \hat S_r \left( {v'} \right) = \mathop \bigwedge \limits_{{\left| u \right|} \le v_r^{\max } } \left( {\hat S_{r - 1} \left( v \right) + \lambda \left| u \right|} \right) = \hfill \\
\underbrace {\mathop \bigwedge \limits_{0 \le u } \hspace{-0.055 in}\left( \hspace{-0.02 in}{\hat S_{r - 1} \left( v \right) + \lambda u}\hspace{-0.02 in} \right)}_{\hat S_r^ + \left( {v'} \right)} \wedge \hspace{-0.05 in}\underbrace {\mathop \bigwedge \limits_{u < 0 } \hspace{-0.05 in}\left(\hspace{-0.02 in} {\hat S_{r - 1} \left( v \right) - \lambda u} \hspace{-0.02 in}\right)}_{\hat S_r^ - \left( {v'} \right)}\hspace{-0.02 in}. \hfill \\ 
\end{array}  
\label{eq:L_dec2}\end{equation}
The sums $\hat S_r^ - \left( {v'} \right)$ and $\hat S_r^ + \left( {v'} \right)$ can be calculated recursively 
\begin{equation}
\begin{array}{l}
 \hat S_r^ + \left( {v'} \right) \hspace{0.23 in}= \hspace{0.1 in}\mathop \bigwedge \limits_{v \le v'} \left( {\hat S_{r - 1} \left( v \right) + \lambda v' - \lambda v} \right), \hfill \\
 \hat S_r^ + \left( {v' + 1} \right) = \hspace{0.05 in}\mathop \bigwedge \limits_{v \le v' + 1} \hspace{-0.045 in} \left( {\hat S_{r - 1} \left( v \right) + \lambda v' - \lambda v} \right) + \lambda \hfill \\
 \hspace{0.73 in} = \hspace{0.1 in} \left( {\hat S_r^ + \left( {v'} \right) + \lambda } \right) \wedge \hat S_{r - 1} \left( {v' + 1} \right), \hfill \\ \hfill \\
 \hat S_r^ + \left( 1 \right) \hspace{0.33 in} = \hspace{0.1 in} \hat S_{r - 1} \left( 1 \right), \hfill \\  
\end{array}  
\label{eq:L_dec3}\end{equation}
and
\begin{equation}
\begin{array}{l}
 \hat S_r^ - \left( {v'} \right) \hspace{0.23 in}= \hspace{0.1 in} \mathop \bigwedge \limits_{v' < v} \left( {\hat S_{r - 1} \left( v \right) - \lambda v' + \lambda v} \right), \hfill \\
 \hat S_r^ - \left( {v' - 1} \right) = \hspace{0.05 in} \mathop \bigwedge \limits_{v' < v + 1} \hspace{-0.045 in} \left( {\hat S_{r - 1} \left( v \right) - \lambda v' + \lambda v} \right) + \lambda \hfill \\
 \hspace{0.73 in} = \hspace{0.1 in}\left( {\hat S_r^ + \left( {v'} \right) + \lambda } \right) \wedge \hat S_{r - 1} \left( {v' - 1} \right), \hfill \\ \hfill \\
 \hat S_r^ + \left( {v_r^{\max } } \right) \hspace{0.1 in} = \hspace{0.15 in} \infty . \hfill \\ 
\end{array}  
\label{eq:L_dec5}\end{equation}
Finally, we get the fast operator ${\mathbf{L}}_r $
\begin{equation}
\hat S_r \left( {v'} \right) = {\mathbf{L}}_r \left( {\hat S_{r - 1} \left( v \right)} \right) = \hat S_r^ + \left( {v'} \right) \wedge \hat S_r^ - \left( {v'} \right),
\label{eq:L_dec6}\end{equation}
that consists of two pass recursion described in (\ref {eq:L_dec3}) and in (\ref {eq:L_dec5}) and the additional comparison pass in (\ref {eq:L_dec6}).

The process of the fast operator ${\mathbf{L}}_r $ should be done in the whole subdomain $1 \le v_r \left( x \right) \le v_r^{\max } $ and the result of the application of the  ${\mathbf{L}}_r $ have to be placed to the intermediate buffer for the application of the next fast operator ${\mathbf{L}}_{r+1} $ if $R>1 $ like it is illustrated in Fig.~\ref {fg:Cut_wnd_l}. In this illustration we have the 2D disparity space with two instances of the application of the fast operator ${\mathbf{L}}_x $ and  ${\mathbf{L}}_y $. The number of the intermediate buffers in this case is one.  

The sum $S^* $ in step-by-step optimization process (\ref {eq:base}) for the truncated linear prior now is given by 
\begin{equation}
\begin{array}{l}
 S^* \left( {x + 1,{\mathbf{v}}\left( {x + 1} \right)} \right) = {\mathbf{M}}\left( {S\left( {x,{\mathbf{v}}\left( x \right)} \right)   } \right)   = \hfill \\\hfill \\
{\mathbf{ L}}_R \hspace{-0.01 in}\left( \hspace{-0.01 in}{.\hspace{-0.01 in}.\hspace{-0.015 in}.{\mathbf{ L}}_2 \hspace{-0.025 in}\left( {{\mathbf{ L}}_1 \hspace{-0.025 in}\left( {S \hspace{-0.02 in}\left( {x,{\mathbf{v}}\left( x \right)} \hspace{-0.02 in}\right)} \hspace{-0.02 in}\right)}\hspace{-0.02 in} \right)} \hspace{-0.02 in}\right)\hspace{-0.02 in} \wedge \hspace{-0.02 in}\left( {S_{\min }\hspace{-0.025 in} \left( x \right)\hspace{-0.017 in} + \hspace{-0.017 in}\lambda f\hspace{-0.025 in}\left( g \right)} \hspace{-0.02 in}\right)\hspace{-0.015 in}. \hfill \\ 
\end{array} 
\label{eq:L_dec7}\end{equation}
The computational complexity of the fast operator ${\mathbf{L}}_r $ is exactly $O(3)$ per vertex in the DSI space and the speedup of the application of the operator ${\mathbf{L}}_r $ in (\ref {eq:L_dec7}) makes sense only for the threshold value $g>2$. 

Summarizing the RMS approach we have to note that there are three ways of calculating the sum $S^*$ in the step-by-step optimization process (\ref {eq:base}):
\begin{itemize}
\item SFMS - The straightforward minimum search using directly the expression in the formula (\ref {eq:base});
\item GRMS - The general RMS algorithm summarized in (\ref {eq:M_decmp});
\item LRMS - The RMS algorithm for the truncated linear prior summarized in (\ref {eq:L_dec7}). 
\end{itemize}
The above abbreviations are referred in comparison tables in the experimental section of the paper.
\section{DP and EDP Algorithms}
The advantage of the DP approach is shown for the one-dimensional case by comparison with other methods. For example, the BP algorithms need $x^{\max }$ iterations to achieve the exact solution of the problem described in (\ref {eq:One_d}) instead of one in the case of the DP approach application.  
In this section we describe the EDP algorithm that is an extension of the general DP technique to the multidimensional case. However, in the first subsection we have to explain the computational scheme of the standard realization of the DP algorithm and its modification for the one-dimensional case.  
\subsection{	DP Algorithm}
\begin{figure}[!t]
\centering
\includegraphics[width=3.30in]{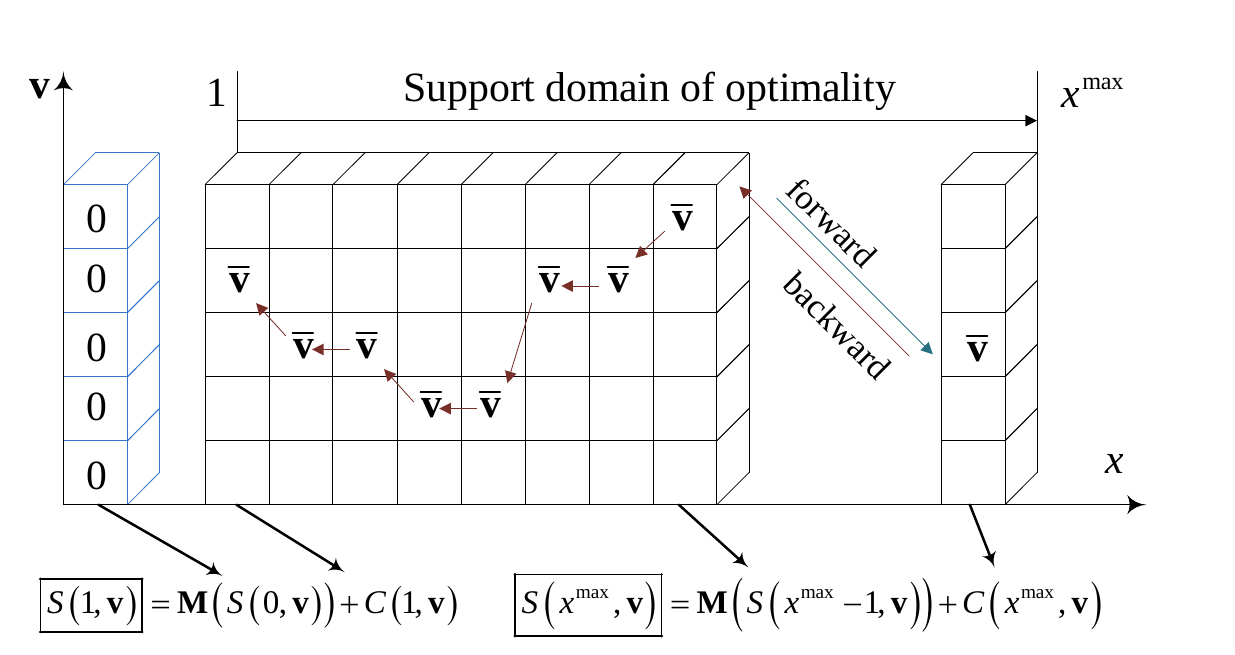}

\centering (a)

\centering
\includegraphics[width=3.30in]{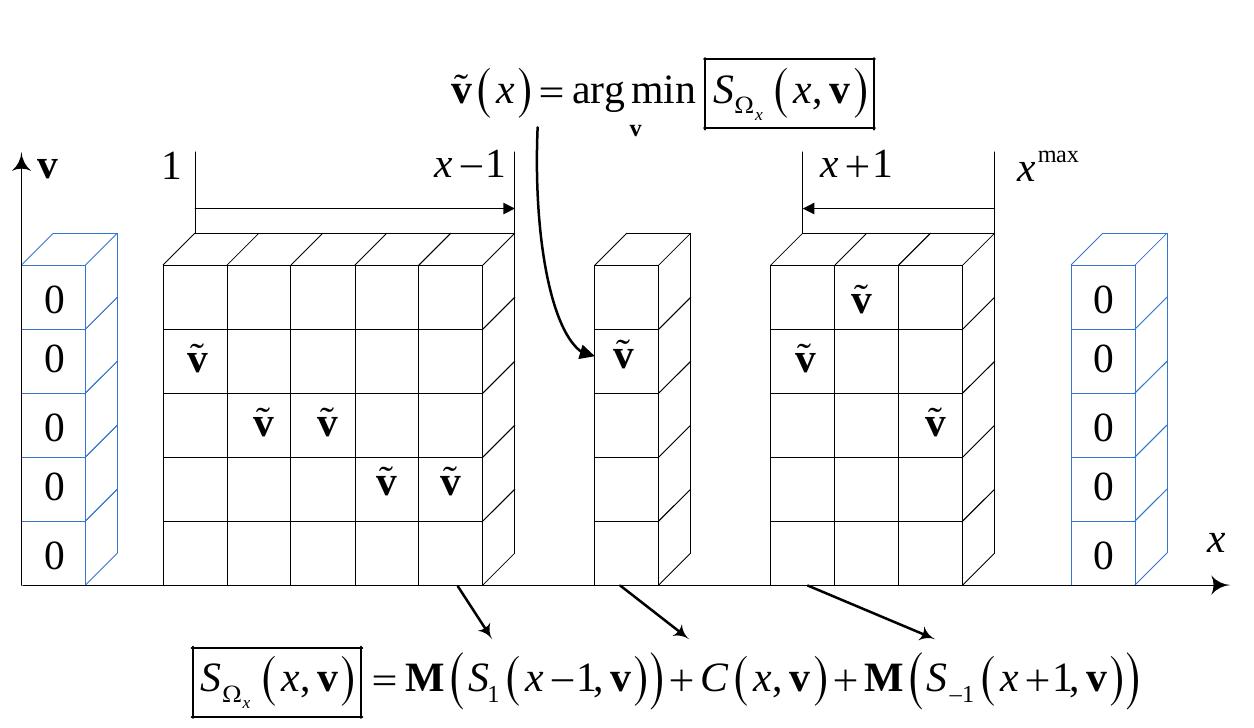}

\centering (b)

\caption{The scheme of the DP agorithm, (a) - standard realization, (b) - modified.}
\label{fg:1D_EDP}
\end{figure}

To make our algorithm more flexible, we omit the initialization step of the classical DP technique in (\ref {eq:base}). Instead, we take the optimal sums in (\ref {eq:base}) equal to zero for all pixels that are not included into the image domain
\begin{equation}
\begin{array}{l}
S\left( {x,{\mathbf{v}}} \right) \equiv 0,\hspace{0.05 in} and \hspace{0.05 in}
 {\mathbf{M}}\left( {S\left( {x,{\mathbf{v}}} \right)} \right) \equiv 0, \forall x \notin \Omega _x, \hfill \\ 
\end{array} 
\label{eq:DP_1D_2}\end{equation}
this condition will hold for all algorithms described in our paper. By using the recurrence relation in (\ref {eq:base}) the minimal value of the objective function in (\ref {eq:One_d}) can be found at the last step of optimization as 
\begin{equation}
S_{\min } \left( {x^{\max } ,{\mathbf{\bar v}}} \right) = \mathop \bigwedge \limits_{\mathbf{v}} S\left( {x^{\max } ,{\mathbf{v}}} \right), 
\label{eq:DP_1D_2}\end{equation}
in which ${\mathbf{\bar v}}$ is the value of the disparity where the sum $S\left( {x^{\max } ,{\mathbf{v}}} \right)
$ reaches its minimum and, in the same time, it is the first value ${\mathbf{\bar v}}\left( {x^{\max } } \right)$
 of the desired solution. Additionally,  it is the starting vertex for the backward optimal path recovery process when the algorithm works in reverse order and recovers a sequence of optimal steps as it is illustrated in Fig.~\ref {fg:1D_EDP}(a). Where the forward steps include the storage of the optimal choice indexes, which are used in the backward steps. 

The sum $S\left( {x^{\max } } \right)$ is the marginal function for the support domain of optimality $\left\{ {1,...,x,x + 1,...,x^{\max } } \right\}$. If we change the direction of the recurrence (starting from the vertex $x^{\max } $ to the vertex 1) the marginal function for the support domain of optimality $\left\{ {x^{\max } ,...,x,x - 1,...,1} \right\}$ will be the $S\left( 1 \right)$. Let the sum $S_{1} \left( {x - 1} \right)$ be the marginal function for the support domain of optimality $\left\{ {1,2,...,x - 1} \right\}$ and the sum $S_{ - 1} \left( {x + 1} \right)$ be the marginal function for the support domain of optimality $\left\{ {x^{\max } ,x^{\max } - 1,...,x + 1} \right\}$, then the marginal function $S_{\Omega _x } \left( x \right)$ for the full domain $x \in \Omega _x $ is 
\begin{equation}
S_{\Omega _x } \hspace{-0.042 in}\left( \hspace{-0.018 in}{x,\hspace{-0.033 in}{\mathbf{v}}} \hspace{-0.017 in}\right)\hspace{-0.023 in} = \hspace{-0.023 in}{\mathbf{M}}\hspace{-0.027 in}\left( \hspace{-0.015 in}{S_{1} \hspace{-0.035 in} \left( {x\hspace{-0.023 in} - \hspace{-0.023 in}1,\hspace{-0.033 in}{\mathbf{v}}} \right)} \hspace{-0.017 in}\right)\hspace{-0.023 in} + \hspace{-0.023 in}C\hspace{-0.025 in}\left(\hspace{-0.015 in} {x,\hspace{-0.015 in}{\mathbf{v}}} \hspace{-0.0125 in}\right) \hspace{-0.02 in} + \hspace{-0.02 in}{\mathbf{M}}\hspace{-0.027 in}\left(\hspace{-0.017 in} {S_{ - 1} \hspace{-0.035 in}\left( {x\hspace{-0.023 in} + \hspace{-0.023 in}1,\hspace{-0.033 in}{\mathbf{v}}} \right)} \hspace{-0.017 in}\right)\hspace{-0.017 in}.
\label{eq:DP_1D_3}\end{equation}
Then, the desired solution can be obtained in each vertex $x$ by
\begin{equation}
{\mathbf{\tilde v}}\left( x \right) = \mathop {\arg \min }\limits_{\mathbf{v}} \left( {S_{\Omega _x } \left( {x,{\mathbf{v}}} \right)} \right).
\label{eq:DP_1D_5}\end{equation}
It was proven in \cite{bib:Mozerov2} that ${\mathbf{\tilde v}}$ is an exact solution, ${\mathbf{\tilde v}}\left( x \right) = {\mathbf{\bar v}}\left( x \right)$, if this optimal solution is unique and an approximation otherwise. The process of the modified DP algorithm is illustrated in Fig.~\ref {fg:1D_EDP}(b).  

The modification of the DP algorithm in (\ref {eq:DP_1D_3}) and (\ref {eq:DP_1D_5}) are not useful for the one-dimensional case, because the computational complexity increases twice in this case, however, the idea to divide the image domain into  a set of support subdomains becomes a significant advantage for the multidimensional case.

\subsection{	EDP Algorithm}
\begin{figure}[!t]
\centering
\includegraphics[width=3.30in]{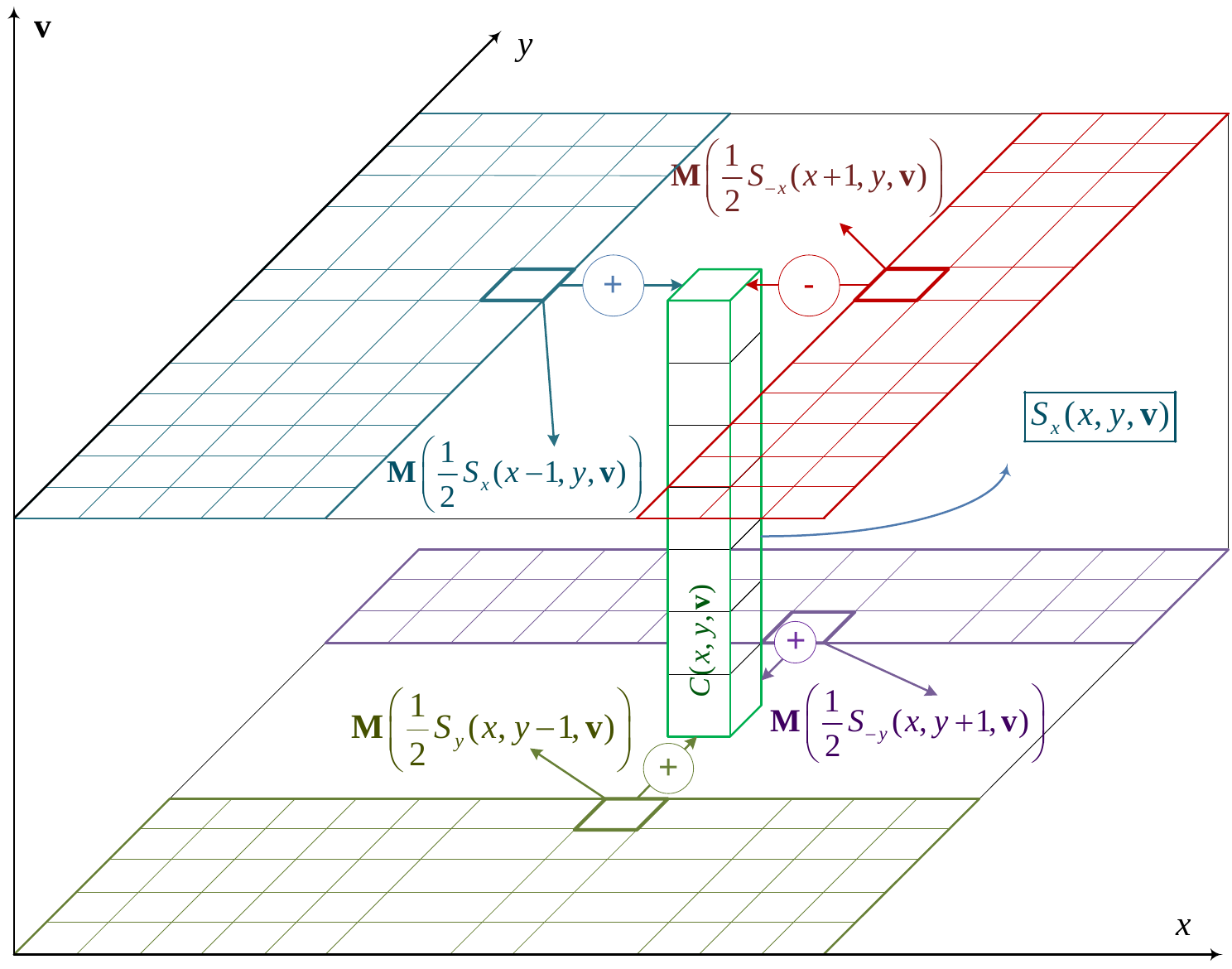}




\caption{ Recursion of an optimal sum $S_x \left( {x,y,{\mathbf{v}}} \right)$.}
\label{fg:2D_EDP}
\end{figure}
The main goal of the EDP algorithm is to estimate marginal function $S_{\mathbf{\Omega }} \left( {{\mathbf{x}},{\mathbf{v}}} \right)$. Unfortunately, the exact estimation of the mentioned marginal function is not possible for the case of the multidimensional variable $\mathbf{x}$, like it can be done for the one-dimensional case and was described in the previous subsection. Nevertheless, an excellent approximation can be done, using the similar approach.   
Let $S_k \left( {x_1 ,...,x_{\left| k \right|} ,..x_D ,{\mathbf{v}}} \right)$
be a sum of optimal costs that covers a support domain ${S_k}\left( {{\bf{x}},{\bf{v}}} \right) \to {\bf{x'}} \in \left\{ {\forall \Omega ,{\mathop{\rm sgn}} \left( k \right){{x'}_{\left| k \right|}} \le {\mathop{\rm sgn}} \left( k \right){x_{\left| k \right|}}} \right\}$
 for all the $k \in K = \left\{ { \pm 1, \pm 2,..., \pm D} \right\}$
 as illustrated in Fig.~\ref {fg:2D_EDP}, where $D=2$ and the number of types of the optimal sums $S_k $
 is 4. In the general multidimensional case, this number is $2D$, which is twice as much as the dimensionality of the image space. We also introduce an auxiliary shortcut  
for the image space variable 
${{\bf{x}}_k} = \left\{ {{x_1},{x_2},...,{x_{\left| k \right|}} - {\mathop{\rm sgn}} \left( k \right),...,{x_D}} \right\}$, 
then the marginal sum $S_{\mathbf{\Omega }} \left( {{\mathbf{x}},{\mathbf{v}}} \right)$ of the whole image space domain ${\mathbf{\Omega }}$
is getting
\begin{equation}
S_{\mathbf{\Omega }} \left( {{\mathbf{x}},{\mathbf{v}}} \right) = \sum\limits_{k \in K} {{\mathbf{M}}\left( {\frac{1}
{D}S_k \left( {{\mathbf{x}}_k ,{\mathbf{v}}} \right)} \right)} ,
\label{eq:DP_2D_1}\end{equation}
 where the minimum search operator ${\mathbf{M}}$
 has the same meaning as in (\ref {eq:base}) and (\ref {eq:M_strf}). The normalization factor $\frac{1}
{D}$ in (\ref {eq:DP_2D_1}) indicates that the integral number of the image space vertex ${\mathbf{x}}$, which is covered by the $2D$ support domains of the sums $S_k \left( {{\mathbf{x}},{\mathbf{v}}} \right)$ in (\ref {eq:DP_2D_1}), is $D$ times of the $\left| {\mathbf{\Omega }} \right|$.
The estimation of the marginal function $S_{\mathbf{\Omega }} \left( {{\mathbf{x}},{\mathbf{v}}} \right)$ in (\ref {eq:DP_2D_1}) supposes that all $k$ optimal sums $S_k \left( {{\mathbf{x}},{\mathbf{v}}}\right)$ are yet calculated. 
The sums $S_k \left( {{\mathbf{x}},{\mathbf{v}}} \right)$, in turn, can be calculated recursively 
\begin{equation}
\begin{array}{l}
 S_{k'} \left( {{\mathbf{x}},{\mathbf{v}}} \right) = \sum\limits_{k \in K/ - k'} {{\mathbf{M}}\left( {\frac{1}
{D}S_k \left( {{\mathbf{x}}_k ,{\mathbf{v}}} \right)} \right)} \hfill \\
  - {\mathbf{M}}\left( {\frac{1}
{D}S_{ - k'} \left( {{\mathbf{x}}_{ - k'} ,{\mathbf{v}}} \right)} \right). \hfill \\ 
\end{array}
\label{eq:DP_2D_base}\end{equation}
The process of this stage of the EDP algorithm is illustrated in Fig.~\ref {fg:2D_EDP}, where the $S_x \left( {x,y,{\mathbf{v}}} \right)$ is one of the four possible optimal sums to be calculated. 
 The approximation of the desired solution in (\ref {eq:GE_objective_function}) now is
\begin{equation}
{\mathbf{\tilde v}}\left( {\mathbf{x}} \right) = \mathop {\arg \min }\limits_{\mathbf{v}} \left( {S_{\mathbf{\Omega }} \left( {{\mathbf{x}},{\mathbf{v}}} \right)} \right).
\label{eq:DP_2D_2}\end{equation}
Note that the solution in (\ref {eq:DP_2D_2}) has not the same meaning than the optimal solution in (\ref {eq:DP_1D_5}) because in the case of the EDP the marginals are only estimated, unlike in (\ref {eq:DP_1D_5}) where they are exact. 
The equation (\ref {eq:DP_2D_base}) is the base formula of the EDP algorithm and the case of $S_x \left( {{\mathbf{x}},{\mathbf{v}}} \right)$ (or $S_1 \left( {{\mathbf{x}},{\mathbf{v}}} \right)$) calculation in the 2D image space is illustrated in Fig.~\ref {fg:2D_EDP}, where the sum $S_x $ in the vertex $(x,y)$ is calculated recursively using the previously calculated sums $\left\{ {S_x ,S_y ,S_y ,S_{ - y} } \right\}$ in the vertices $\left\{ {\left( {x - 1,y} \right),\left( {x,y - 1} \right),\left( {x + 1,y} \right),\left( {x,y + 1} \right)} \right\}$ respectively. The both formulas (\ref {eq:DP_2D_1}) and (\ref {eq:DP_2D_base})  were not strictly derived and, in fact, are a heuristic.  The main reason we chose this forms instead of something else is based on the cardinality preservation for the support domains of the sums $S_k \left( {{\mathbf{x}},{\mathbf{v}}} \right)$. Let $\sigma \left( {S_k \left( {\mathbf{x}} \right)} \right)$ be the cardinality of the support domain of the sum $S_k \left( {\mathbf{x}} \right)$ (for example $\sigma \left( {S_{\mathbf{\Omega }} \left( {\mathbf{x}} \right)} \right) = N$), then 
\begin{equation}
\sigma \left( {S_{\mathbf{\Omega }} \left( {\mathbf{x}} \right)} \right) \simeq \frac{1}
{D}\sum\limits_{k \in K} {\sigma \left( {S_k \left( {{\mathbf{x}}_k } \right)} \right)},
\label{eq:DP_2D_1_1}\end{equation}
and
\begin{equation}
\begin{array}{l}
 \sigma \left( {S_{k'} \left( {\mathbf{x}} \right)} \right) \simeq \frac{1}
{D}\sum\limits_{k \in K/ - k'} {\sigma \left( {S_k \left( {{\mathbf{x}}_k } \right)} \right)} \hfill \\
  - \frac{1}
{D}\sigma \left( {S_{ - k'} \left( {{\mathbf{x}}_{ - k'} } \right)} \right). \hfill \\  
\end{array}
\label{eq:DP_2D_base_1}\end{equation}
Thus, the comparison of (\ref {eq:DP_2D_1}) and (\ref {eq:DP_2D_base}) with (\ref {eq:DP_2D_1_1}) and (\ref {eq:DP_2D_base_1}) explains why we chose the mentioned heuristic.
  
If the calculation of the sums $S_k $ in (\ref {eq:DP_2D_base}) is organized by iteration:
\begin{equation}
\begin{array}{l}
 S_{k'}^{\tau + 1} \left( {{\mathbf{x}},{\mathbf{v}}} \right) = \sum\limits_{k \in K/ - k'} {{\mathbf{M}}\left( {\frac{1}
{D}S_k^\tau \left( {{\mathbf{x}}_k ,{\mathbf{v}}} \right)} \right)} \hfill \\
  - {\mathbf{M}}\left( {\frac{1}
{D}S_{ - k'}^\tau \left( {{\mathbf{x}}_{ - k'} ,{\mathbf{v}}} \right)} \right). \hfill \\  
\end{array}
\label{eq:DP_2D_base_2}\end{equation}
Then the result of the energy minimization will be the same as in the loopy belief propagation (LBP) approach \cite{bib:LBP1, bib:LBP2}, which is reported in the evaluation paper \cite{bib:Szeliski}: it shows a slow convergence to 110-125\% energy level of the GC expansion algorithm.

Therefore, to improve LBP results our algorithm has to inherit the recursive property of the DP approach. Thus, the EDP algorithm is essentially recursive instead of iterative. Of course, we can also use several iterations to improve the level of the energy minimization; however, in this case our algorithm never switches from $\tau $ to $\tau + 1$ as it is necessary for iterative algorithm, like for instance in (\ref {eq:DP_2D_base_2}). Note that recursion and iteration are sometimes considered equivalent, but here we consider iteration as a data transformation from one instance of the whole data set to another instance. In contrast, recursion is considered as a data transformation from one node of the data set to a neighbor. Ideally, in the case of recursion, an impact of one node value reaches any other node of the data set within one iteration, like it is in the case of DP, for instance.  In such a way, the EDP recursion gives advantage over pure BP iterations, where information from one node reaches another after several iterations. 

Let us consider a vertex scanning procedure (with different scanning trees) of a rectangular image in $2D$. The simplest and the most popular scanning procedure consist of two embedded incremental loops: 
one intrinsic loop is a set of incremental steps in the $x$ coordinate of the image plane 
${\rm{inc}}\left( x \right) = \left( {1 \to 2 \to ... \to x \to x + 1 \to ... \to {x^{\max }}} \right)$ 
with a fixed value of the $y$; and another extrinsic loop is a set of incremental steps in the $y$ coordinate of the image plane 
${\mathop{\rm inc}\nolimits} \left( y \right) = \left( {1 \to ... \to y \to ... \to {y^{\max }}} \right)$. The both loops can be also organized in the decrementing order, for example for the $x$: 
${\mathop{\rm dec}\nolimits} \left( x \right) = \left( {{x^{\max }} \to ... \to x \to ... \to 1} \right)$. Finally there are four simple scanning procedures in 2D with four different pairs of starting-ending vertices:  
\begin{itemize}
\item ${{\mathop{\rm P}\nolimits} _1} = {\mathop{\rm inc}\nolimits} \left( {{\mathop{\rm inc}\nolimits} \left( x \right),y} \right) = \left( {\left( {1,1} \right) \to \left( {{x^{\max }},{y^{\max }}} \right)} \right)$;
\item ${{\mathop{\rm P}\nolimits} _2} = {\mathop{\rm inc}\nolimits} \left( {{\mathop{\rm dec}\nolimits} \left( x \right),y} \right) = \left( {\left( {{x^{\max }},1} \right) \to \left( {1,{y^{\max }}} \right)} \right)$;
\item ${{\mathop{\rm P}\nolimits} _3} = {\mathop{\rm dec}\nolimits} \left( {{\mathop{\rm inc}\nolimits} \left( x \right),y} \right) = \left( {\left( {1,{y^{\max }}} \right) \to \left( {{x^{\max }},1} \right)} \right)$;
\item ${{\mathop{\rm P}\nolimits} _4} = {\mathop{\rm dec}\nolimits} \left( {{\mathop{\rm dec}\nolimits} \left( x \right),y} \right) = \left( {\left( {{x^{\max }},{y^{\max }}} \right) \to \left( {1,1} \right)} \right)$.
\end{itemize}
If the scanning order of the image is chosen as, for instance ${{\mathop{\rm P}\nolimits} _1} $, for the sums $S_k$
calculation in (\ref {eq:DP_2D_base}), then only the $S_x =S_1$ (see Fig.~\ref {fg:2D_EDP})
 and the $S_y =S_2$
 are calculated recursively.
Thus, the main part of the EDP algorithm consists of the four passes of the sums $S_k $
calculation:
\begin{itemize}
\item Calculate $S_x $ and $S_y $ with  (\ref {eq:DP_2D_base}) under the scanning procedure ${{\mathop{\rm P}\nolimits} _1}$;
\item Calculate $S_{ - x} $ and $S_y $ with  (\ref {eq:DP_2D_base}) under the scanning procedure ${{\mathop{\rm P}\nolimits} _2}$;
\item Calculate $S_x $ and $S_{ - y} $ with  (\ref {eq:DP_2D_base}) under the scanning procedure ${{\mathop{\rm P}\nolimits} _3}$;
\item Calculate $S_{ - x} $ and $S_{ - y} $  (\ref {eq:DP_2D_base}) under the scanning procedure ${{\mathop{\rm P}\nolimits} _4}$.
\end{itemize}
Let us unify all previously described four image scan passes of (\ref {eq:DP_2D_base}) into one iteration step $S_k \left( {{\mathbf{x}},{\mathbf{v}}} \right) = \mathbb{P}\left( {S_k \left( {{\mathbf{x}},{\mathbf{v}}} \right)} \right)$.
Then the summarized EDP algorithm consists of the next steps:
\begin{algorithm}
\caption{EDP for Energy Minimization}
\label{alg1}
\begin{algorithmic}[1]
\STATE Form the DSI by calculating the cost values $C\left( {{\mathbf{x}},{\mathbf{v}}} \right)$ with (\ref {eq:Cost_value}).
\STATE Initialize the function $\bar G({\mathbf{v}}({\mathbf{x}}))$ in (\ref {eq:Smooth_value_t}) for the minimum search operator ${\mathbf{M}}$ in (\ref {eq:M_strf}) with experimental parameters $\lambda $, $g$ and a chosen function $f$.
\STATE Perform $j$ iterations of the procedure $\mathbb{P}$ with (\ref {eq:DP_2D_base}), where $j>0$.
\STATE Calculate the marginal sums $S_{\mathbf{\Omega }} $ with (\ref {eq:DP_2D_1}).
\STATE Obtain the approximation ${\mathbf{\tilde v}}$ of the desired function of the disparity map with (\ref {eq:DP_2D_2}).
\end{algorithmic}
\end{algorithm}

\section{	Experimental Results}
In this section we consider two different aspects of the technique that is proposed in our paper. First, we show the advantages of using the DP approach in discrete energy minimization with comparative results obtained by our EDP algorithm and the GC expansion. Then, in the second subsection, the computational speedup that can be obtained by application of our RMS algorithm is discussed on the base of the previously presented experimental setup.  

\subsection{	Energy Minimization}

In stereo, the most successful methods (see Middlebury stereo evaluation table), use energy minimization approach, such as GC and LBP. In other words, solving the problem in (\ref {eq:GE_objective_function}) is the core of the contemporary state-of-art stereo matching techniques. On the other hand, to reach the top-ten results method should include a cascade of different image processing techniques , for example in \cite{bib:Yang}, where  energy minimization for the MAP matching model is the core algorithm. So, authors in \cite{bib:Yang} use hierarchical belief propagation to solve energy minimization problem, and in principle this core algorithm can be replaced by our EDP approach to improve energy minimization accuracy and speedup the calculation process. We propose to compare our EDP approximation with one of the most excellent energy minimization algorithm: GC expansion that was described in \cite{bib:Boykov}. There are two main reasons of our choice. Firstly, the expansion algorithm is relatively fast (among energy minimization algorithms, which can achieve the same approximation level) and allows to obtain excellent approximate solution in the energy minimization problem. Also, we use an open C++ code for the expansion algorithm (taken from the Middlebury data set) to make comparisons with our EDP approximation. However, we also provide a qualitative comparison with the energy minimization approach proposed in \cite{bib:Kumar}, which is reported to be the best approximation.  

This subsection includes four experimental setups: two in stereo and two in motion with the 2D disparity space. Usually, stereo matching methods include additional cues in the smoothness term of (\ref {eq:GE_objective_function}), pre- and post-processing \cite{bib:Scharstein} to improve the final result. In our experiments we omit these algorithm stages, because it helps explaining the main idea of the paper more clearly.  

In our experiments only two kinds of the prior in (\ref {eq:Smooth_value_t}) with $f\left( {\left| {\mathbf{v}} \right|} \right) = \left| {\mathbf{v}} \right|^{l_1 } $ are used: the case of the truncated linear $l_1=1$ and the case of the truncated squared $l_1=2$ priors. Two kinds of the cost in (\ref {eq:Cost_value}) also the case of the truncated linear $l_2=1$ and the case of the truncated squared $l_2=2$ cost calculation with $C_{\max } = \left( {100} \right)^{l_2 } $.  
The parameter $\lambda $ of the prior function in (\ref {eq:Smooth_value_t}) in all our experiments is taken proportional to the mean value $\left\langle {C\left( {{\mathbf{x}},{\mathbf{v}}} \right)} \right\rangle $ of the DSI cost
\begin{equation}
\lambda = \left\lfloor {\frac{{ l_2 \left\langle {C\left( {{\mathbf{x}},{\mathbf{v}}} \right)} \right\rangle }}
{{l_1 \left| g \right|^{l_1 } }}} \right\rfloor ,
\label{eq:Cost_3}\end{equation}
To make our comparison with GC expansion feasible we rounded the $\lambda $ by the floor function $\left\lfloor {} \right\rfloor $ in (\ref{eq:Cost_3}). It means that all the values of the input prior matrix for the GC expansion are integer.
\begin{figure}[!t]
\centering
\includegraphics[width=3.28in]{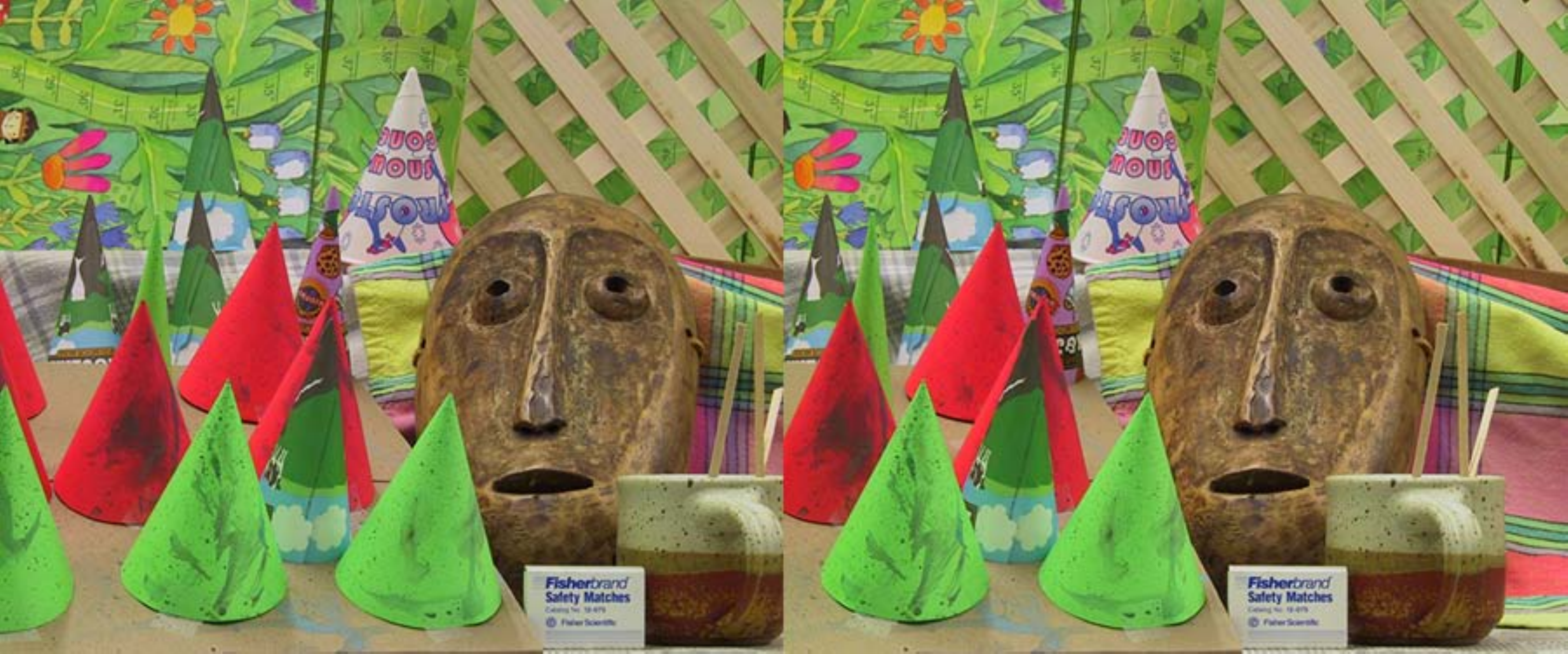}

\centering (a)

\centering
\includegraphics[width=3.28in]{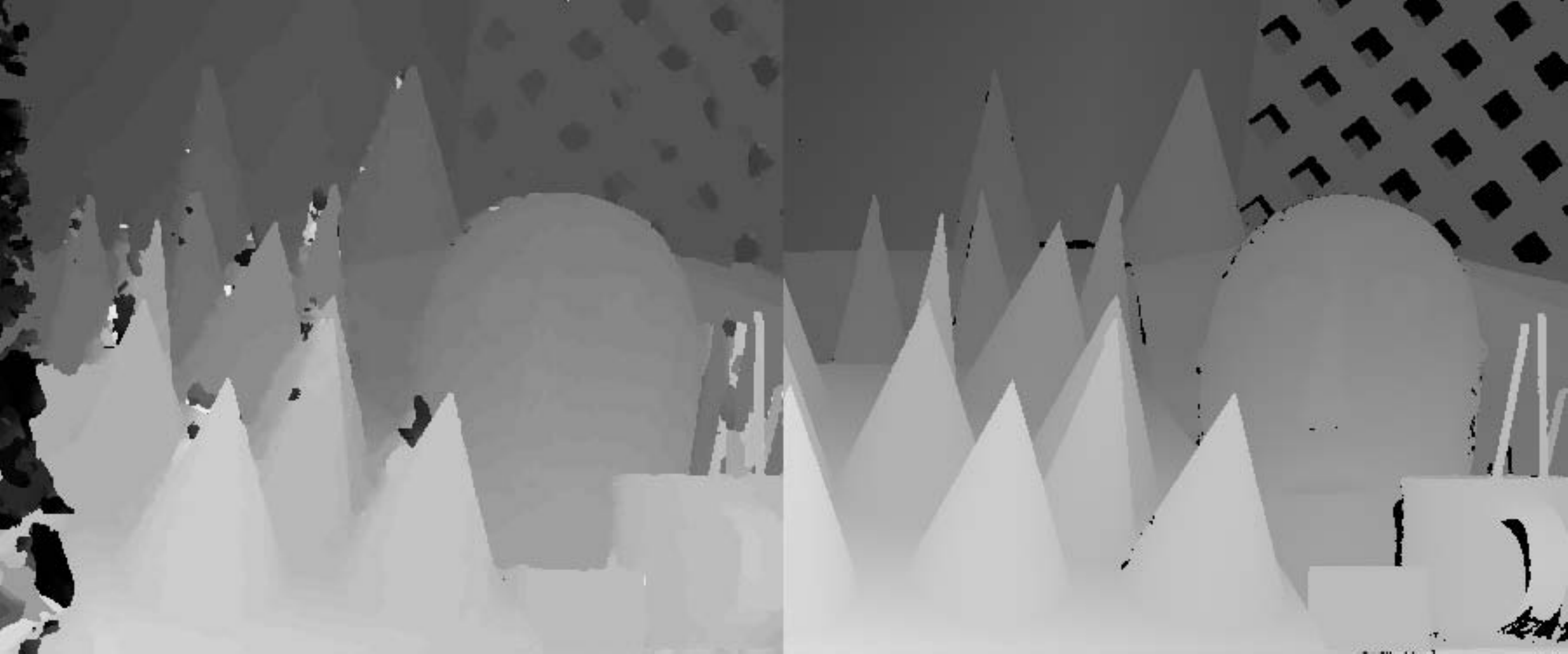}

\centering (b)

\centering
\includegraphics[width=3.23in]{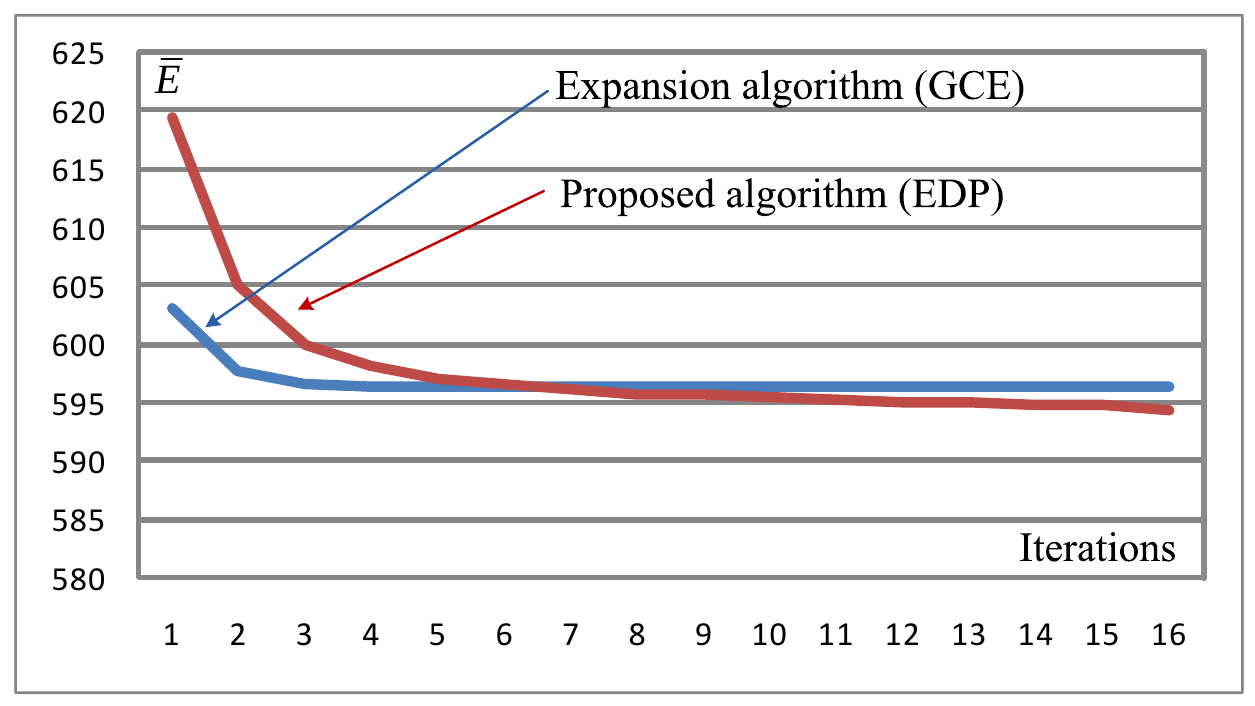}

\centering (c)

\caption{The results of the first experiment, (a) the left and right images of the Cones stereo pair, (b) - the disparity map obtained by the EDP algorithm after first iteration and the ground truth, (c) the comparison diagram of energy values versus iteration for the EDP and the expansion algorithms.}
\label{fg:Exp_1}
\end{figure}

The first experimental setup is based on two stereo images: Cones from the Middlebury data set (the size of the images is 450x375). The maximum disparity in this case is 59 (or 60 labels). The value of power for the prior measure is $l_1 = 1$, which is the case of the truncated linear prior; and the value of power for the cost measure is $l_2 = 2$, which is the case of the truncated squared cost. The truncation threshold is $g=5$. The disparity map obtained by our EDP algorithm after first iteration is shown in Fig.~\ref {fg:Exp_1}(b), where the left part of this image is the resultant disparity map and the right part is the ground truth disparity map.  Fig.~\ref {fg:Exp_1}(a) represents the left and the right images of the Cones stereo pair. 
  
Convergence of the energy value versus number of iteration is shown in Fig.~\ref {fg:Exp_1}(c). We can see that for this particular setup our algorithm surpasses the result of GC expansion algorithm after six iterations. Here and later on we use the per pixel energy value $\bar E = \frac{E} {N}$ instead of the energy value in (\ref {eq:base}), because this value is closer to the values of the mean squared and the mean absolute error criteria. For example, in the diagram in Fig.~\ref {fg:Exp_1}(c) the value of the $\bar E$ is 619.47 and it says that the average value of difference between corresponding color vectors in the left and in the right stereo images is not more than 25.

\begin{figure}[!t]
\centering
\includegraphics[width=3.28in]{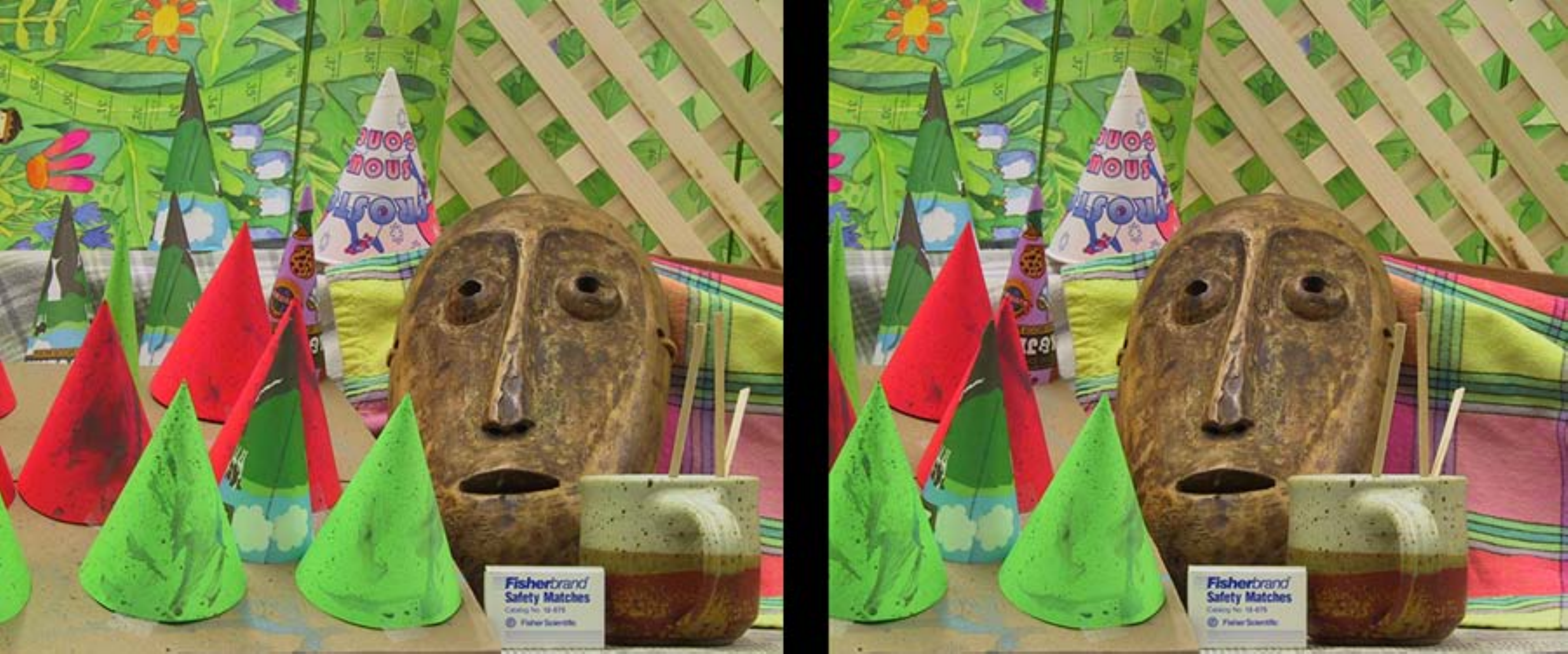}

\centering (a)

\centering
\includegraphics[width=3.28in]{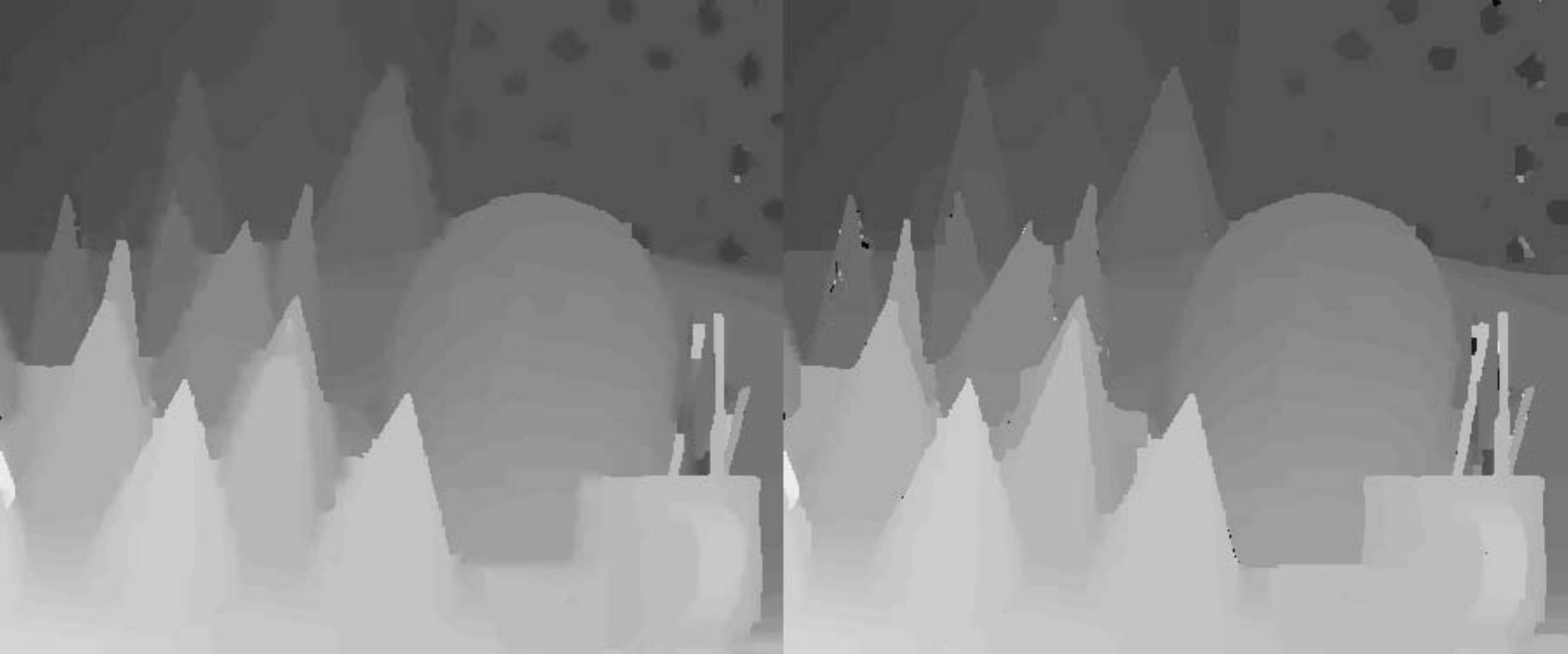}

\centering (b)

\centering
\includegraphics[width=3.23in]{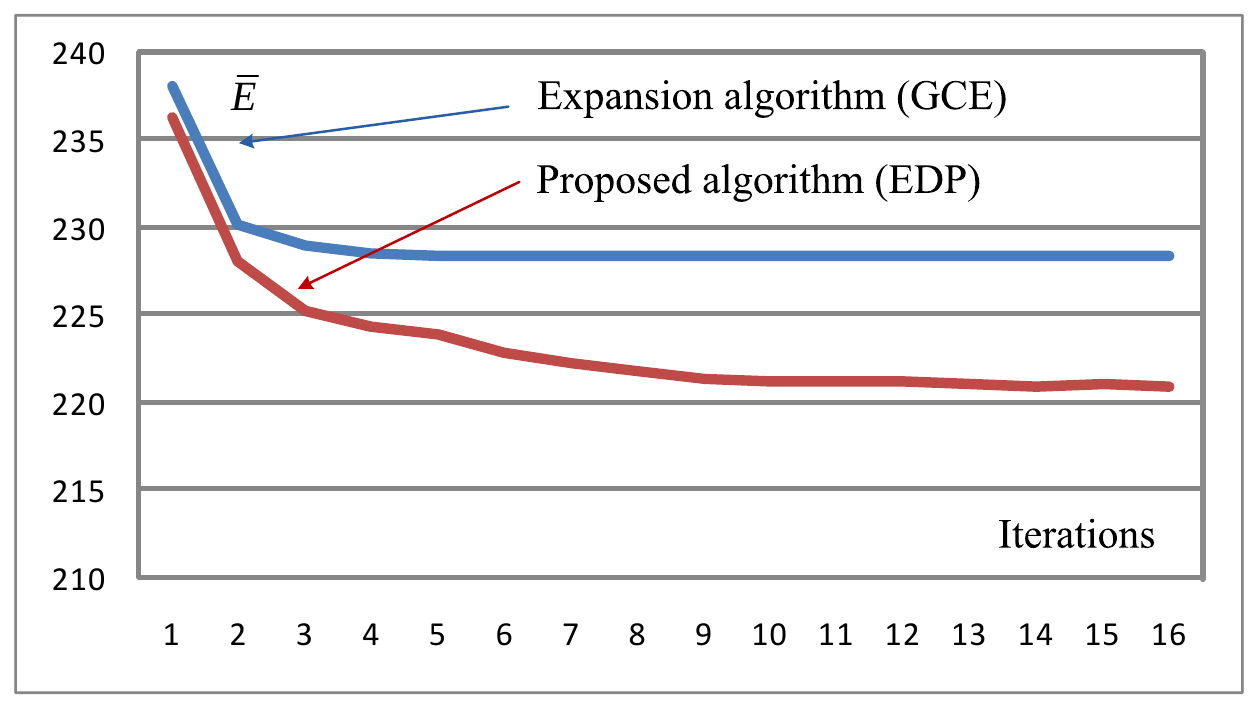}

\centering (c)

\caption{The results of the second experiment, (a) the middle and the right images of the Cones stereo triplet, (b) - the resultant disparity map obtained by the EDP algorithm after 16 iterations for the truncated square and the truncated linear priors, (c) the comparison diagram of energy values versus iteration for the EDP and the expansion algorithms.}
\label{fg:Exp_2}
\end{figure}
The second experimental setup is chosen to demonstrate the possibility of occlusion handling. Such a setup is based on three stereo images: Cones from the Middlebury data set (the size is 450x375). The maximum disparity in this case is 59 (or 60 labels). The value $l_1 = 2$, which is the case of the truncated squared prior; and the value $l_2 = 2$, which is the case of the truncated squared cost. The truncation threshold is $g=3$. The DSI cost in this case is a superposition of two costs: middle-image-to-right and middle-image-to-left
\begin{equation}
C\left( {{\mathbf{x}},{\mathbf{v}}} \right) = C_{mdl \to rgh} \left( {{\mathbf{x}},{\mathbf{v}}} \right) \wedge C_{mdl \to lft} \left( {{\mathbf{x}}, - {\mathbf{v}}} \right),
\label{eq:Cost_cmp}\end{equation}
in which the $C_{mdl \to rgh} \left( {{\mathbf{x}},{\mathbf{v}}} \right)$ and the $C_{mdl \to lft} \left( {{\mathbf{x}}, - {\mathbf{v}}} \right)$ are calculated by (\ref {eq:Cost_value}). The idea of such superposition was proposed in \cite{bib:Mozerov3}. Note than in this case the distance between left and middle camera lenses has to be equal to the distance between right and middle camera lenses, and the axes of all the coordinate systems of the cameras have to be collinear. 
The resultant disparity map obtained by our EDP algorithm after 16 iterations is shown in Fig.~\ref {fg:Exp_2}(b), where the left part of this image is the disparity map obtained with the truncated squared prior and the right part is the disparity map obtained with the truncated linear prior.  Fig.~\ref {fg:Exp_2}(a) represents the left and the right images of the Cones stereo triplet.  We can see that in case of three stereo images   the problem of the disparity map estimation in the occluded regions is practically solved, compare Fig.~\ref {fg:Exp_1}(b) and Fig.~\ref {fg:Exp_2}(b). However, the edges of the disparity map in Fig.~\ref {fg:Exp_2}(b) are smooth due to the squared prior of the experiment setup in comparison with the right part of Fig.~\ref {fg:Exp_2}(b), where the truncated linear prior is used. Convergence of the energy value versus number of iteration is showed in Fig.~\ref {fg:Exp_2}(c). We can see that in this particular prior function our algorithm surpasses the result of GC expansion algorithm from the first iteration.

\begin{figure}[!t]
\centering
\includegraphics[width=3.28in]{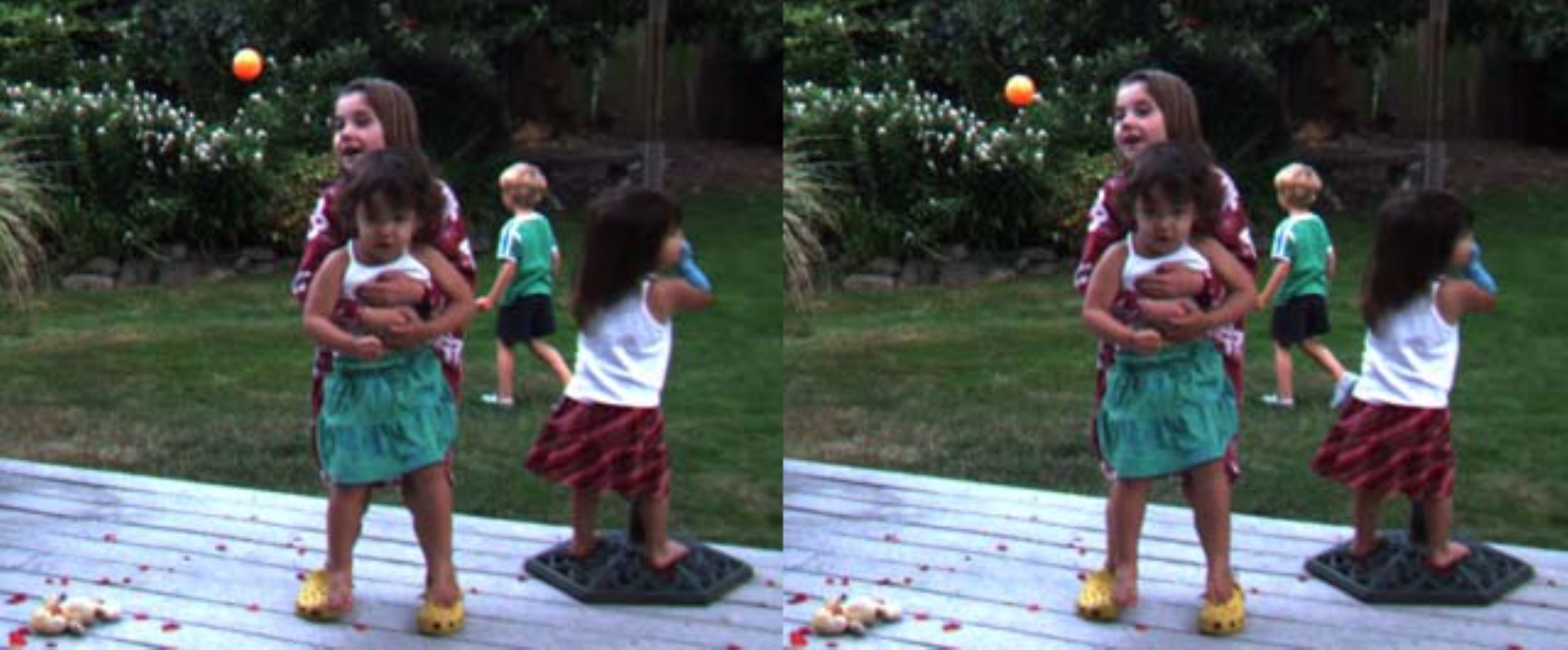}

\centering (a)

\centering
\includegraphics[width=3.28in]{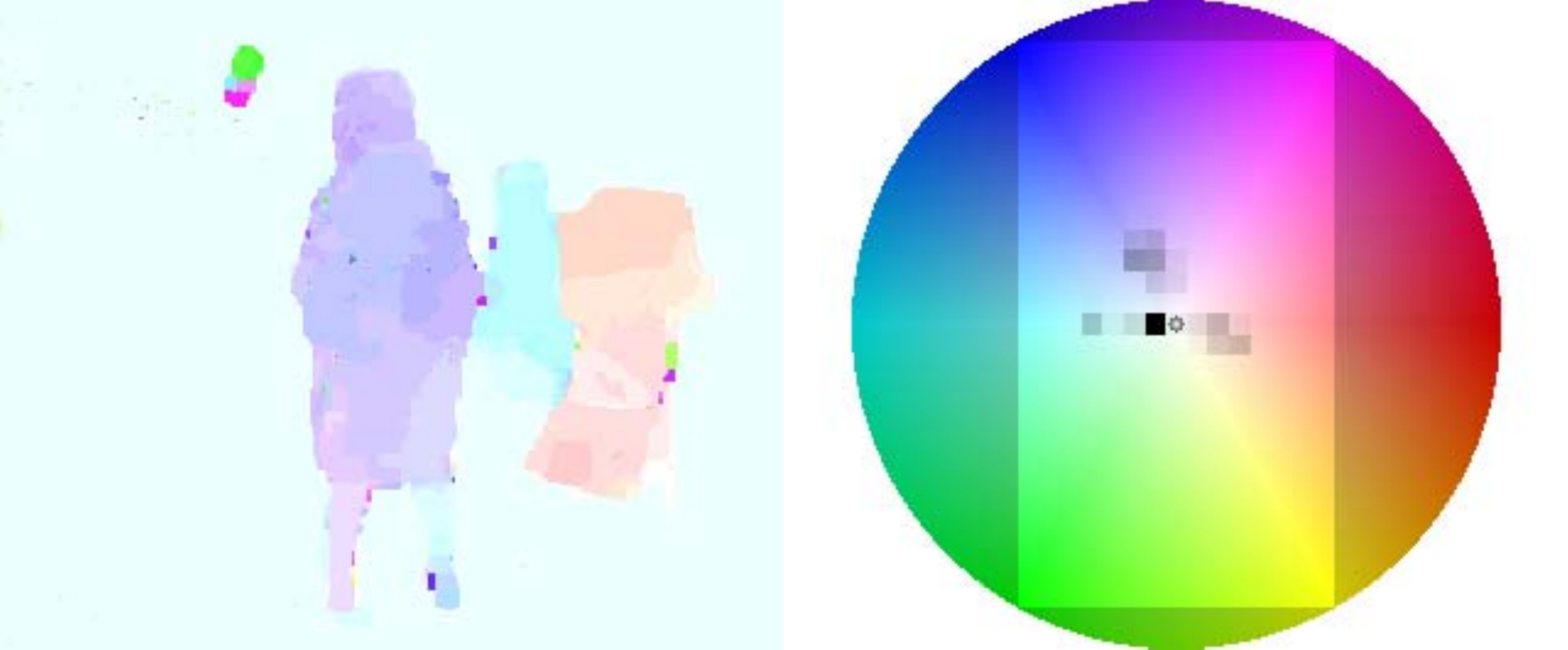}

\centering (b)

\centering
\includegraphics[width=3.23in]{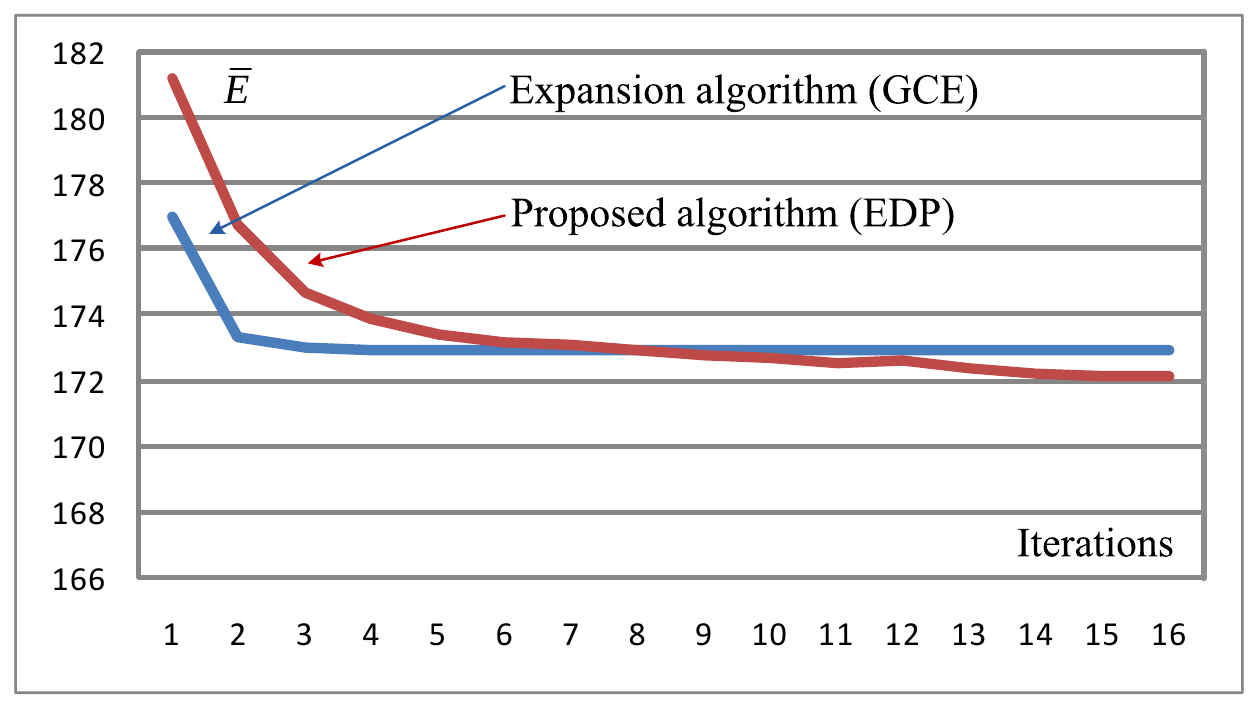}

\centering (c)

\caption{The results of the third experiment, (a) the first and the second images of the Back Yard motion sequence, (b) - the resultant disparity map obtained by the EDP algorithm after the first iteration with the truncated linear prior, (c) the comparison diagram of energy values versus iteration for the EDP and the expansion algorithms.}
\label{fg:Exp_3}
\end{figure}

The third experimental setup is based on two motion images: Back Yard from the Middlebury data set (the size is 320x265). The maximum disparity in this case is $ \pm 7$ in the $v_y$ disparity direction and is $ \pm 13$ in the $v_x$ disparity direction (or 405 labels). The value $l_1 = 1$, which is the case of the truncated linear prior; and the value $l_2 = 2$, which is the case of the truncated squared cost. The truncation threshold $g=3$. The resultant disparity map obtained by our EDP algorithm after first iteration is shown in Fig.~\ref {fg:Exp_3}(b), where Fig.~\ref {fg:Exp_3}(a) is the first and the second images of the Back Yard motion sequence.  The right side of Fig.~\ref {fg:Exp_3}(b) is the hue saturation color map related to the size and orientation of the disparity vector. This map is overlapped by a histogram: the darker pixels in the color map correspond to the more frequent value of the disparity vector in the disparity map.  
Convergence of the energy value versus number of iteration is shown in Fig.~\ref {fg:Exp_3}(c).

\begin{figure}[!t]
\centering
\includegraphics[width=3.28in]{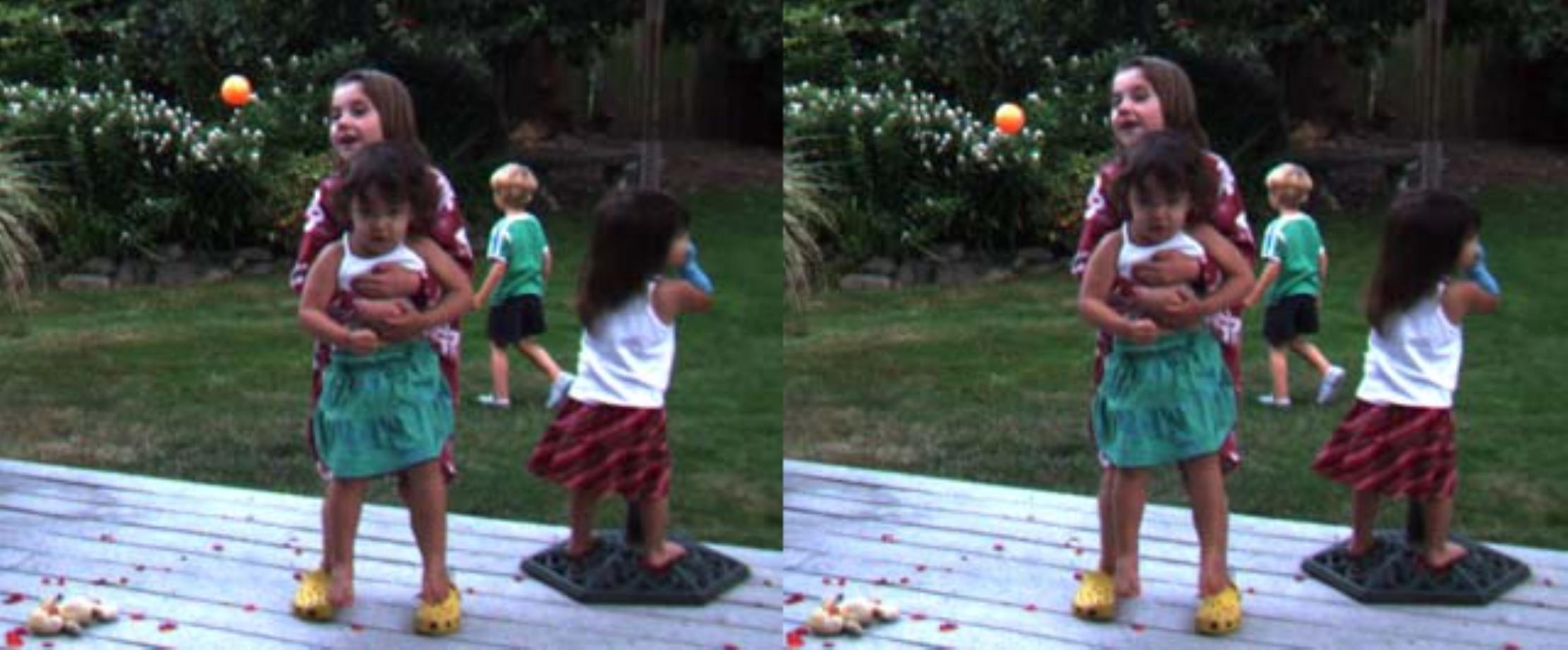}

\centering (a)

\centering
\includegraphics[width=3.28in]{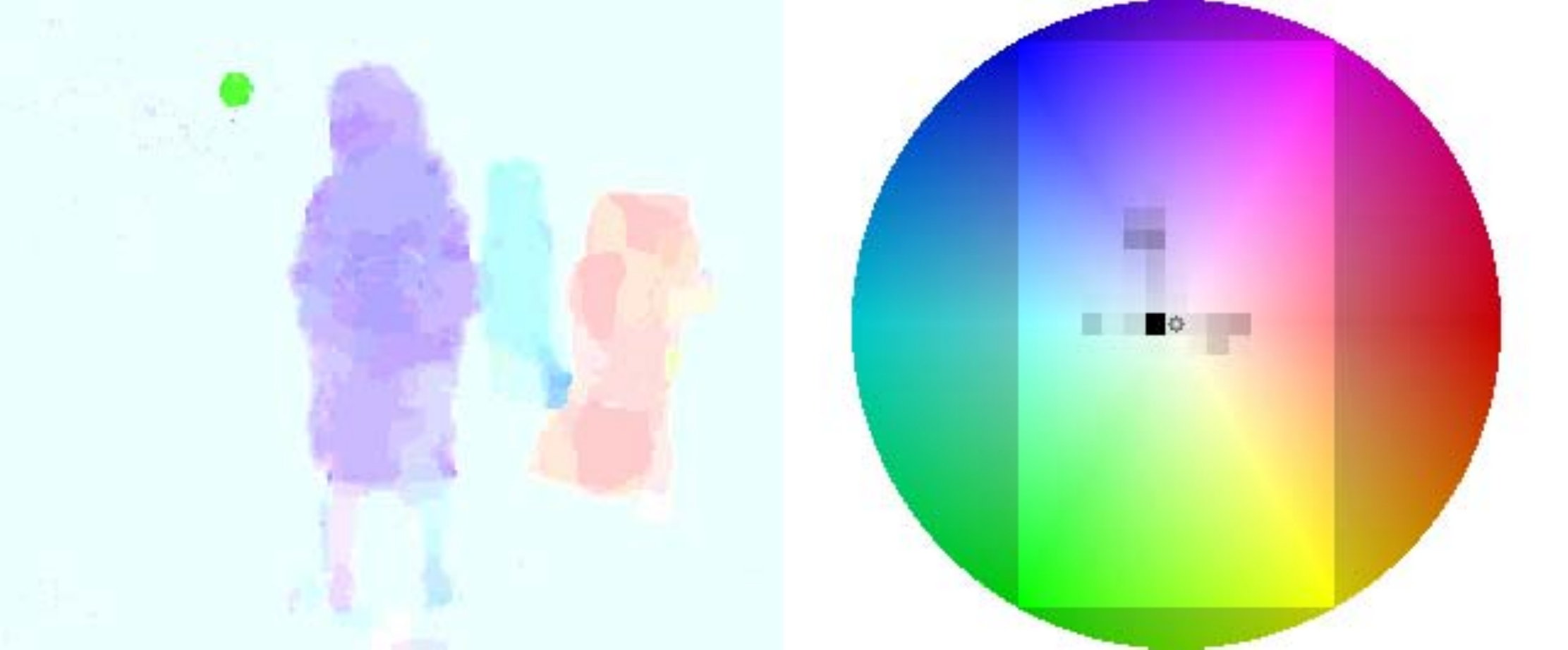}

\centering (b)

\centering
\includegraphics[width=3.23in]{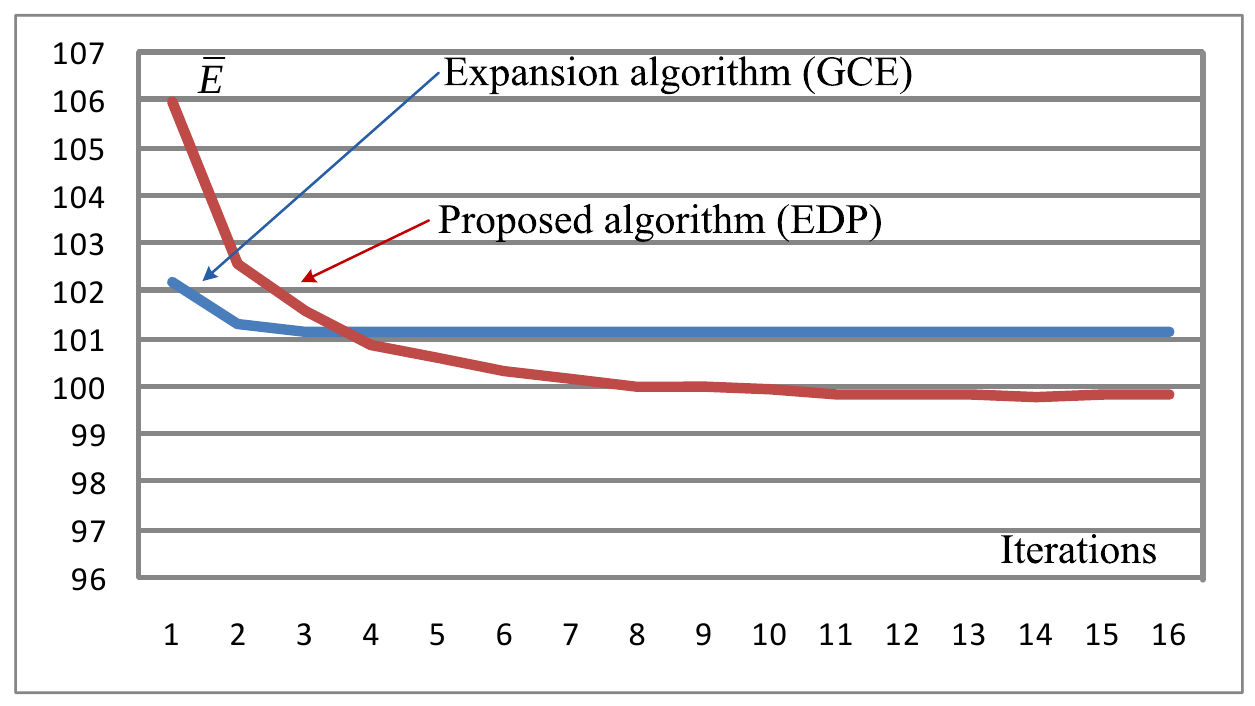}

\centering (c)

\caption{The results of the third experiment, (a) the second and the third images of the Back Yard motion sequence, (b) - the resultant disparity map obtained by the EDP algorithm after 16 iterations with the truncated squared prior, (c) the comparison diagram of energy values versus iteration for the EDP and the expansion algorithms.}
\label{fg:Exp_4}
\end{figure}

The fourth experimental setup is based on three motion images: Back Yard from the Middlebury data set (the size is 320x265). The maximum disparity in this case is $ \pm 7$ in the $v_x$ disparity direction and is $ \pm 13$ in the $v_y$ disparity direction (or 405 labels). The value $l_1 = 2$, which is the case of the truncated squared prior; and the value $l_2 = 2$, which is the case of the truncated squared cost. The truncation threshold $g=3$. The resultant disparity map obtained by our EDP algorithm after 16 iterations is shown in Fig.~\ref {fg:Exp_4}(b), where Fig.~\ref {fg:Exp_4}(a) is the second and the third images of the Back Yard motion sequence.  The DSI cost in this particular case is calculated by (\ref{eq:Cost_cmp}). Here we use the assumption that the forward motion vector (in the 2D disparity space) is approximately equal to the negative backward motion vector. This assumption helps us to overcome the occlusion errors. This is understandable if we compare to resultant disparity maps in Fig.~\ref {fg:Exp_3}(b) and in Fig.~\ref {fg:Exp_4}(b).   
Convergence of the energy value versus number of iteration is shown in Fig.~\ref {fg:Exp_4}(c).

\subsection{	Computational Speedup}
Let us consider more carefully the possible speedup in the case of the implementation of the proposed algorithm general RMS (GRMS) and the linear RMS (LRMS). Theoretically, the slowest straightforward search needs $O(Q)$ operations per vertex in the DSI space where $Q$ is the number of labels. For stereo, the number of labels is $Q= v^{max}$ and for motion is $Q = \left( {2v_x^{\max } - 1} \right)\left( {2v_y^{\max } - 1} \right)$. The application of the GRMS algorithm restrict the number of the needed operations to $R\left( {2g - 1} \right)$ and the application of the LRMS algorithm restrict the number of need operation to $3R$. In Tab.~\ref {table:rlrt} these theoretically derived numbers are given for the four different experimental setups described in Subsection 5.1. The practical realization of the compared algorithms, which are running along, shows nearly the same relative speedup. However, the speedup in the real program realization of the algorithm is not the same; due to additional operations inside the program body (e. g. stack and cycles control operations). In Tab.~\ref {table:base} we give the result of the computer experiments that was done on a standard desktop PC equipped with Core Duo 3.16 GHz CPU 4 GB, where the number of running seconds is per iteration. The time needed per iteration in the case of the expansion algorithm depends on the DSI data and on the number of iterations. Thus, in Tab.~\ref {table:base}, we report the time needed for the first iteration.
\begin{table}
\centering
\caption{
The number of operations per vertex in the DSI space for SFMS, GRMS and LRMS. 
}
\label{table:rlrt}
\begin{tabular}{|c|c|c|c|c|}
\hline
Algorithms & Setup 1 & Setup 2 &
Setup 3 & Setup 4 \\ \hline
SFMS & 60 & 60  & 405  & 405  \\ \hline
GRMS & -  & 9  & - &  10  \\ \hline
LRMS & 3  & -  & 6  & -   \\ \hline

\end{tabular}

\end{table}

Also, our algorithm is running faster than the expansion algorithm, especially in the case when the composite DSI cost calculated with (\ref{eq:Cost_cmp}) is used or the prior dependency is the truncated squared.  

The GCE algorithm was first proposed in  2000. The graph cut community now offers new methods that report better results in terms of energy minimization. See for example approach in \cite{bib:Kumar}. However, if Kumar's method in \cite{bib:Kumar} is compared with  GCE we can see that it is still time consuming: in general this method works 10-25 times slower than the GTE algorithm to reach almost the same result. On the other hand, for the real stereo matching experiments, Kumar's method does not show significant improvement in  terms of energy minimization in comparison with GCE, especially in the case of the truncated linear prior. Indeed,  the difference of the minimum values of two these approaches is less than 0.2\%.  For example, the difference of the minimum values between GCE and LBP can be more than 15\%.   Note that such a difference (0.2\%) in energy minimization has only theoretical impact for reconstruction accuracy. We would like to stress  that our algorithm surpasses the result of GCE by almost the same 0.2\% in the case of the truncated linear prior.  
\begin{table}
\centering
\caption{The speedup of the GRMS and the LRMS versus the SFMS and the GCE algorithms,where the number of running seconds is per iteration.}
\label{table:base} 
\begin{tabular}{|c|c|c|c|c|}
\hline
Algorithms & Setup 1 & Setup 2 &
Setup 3 & Setup 4 \\ \hline
SFMS & 19.1 s & 19.1 s  & 336 s  & 336 s  \\ \hline
GRMS & -  & 3.2 s  & - &  16.6 s  \\ \hline
LRMS & 2.3 s & -  & 10.8 c  & -  \\ \hline
GCE & 4.3 s & 5.1 s & 12.1 s & 15.8 s  \\ \hline
\end{tabular}

\end{table}

\section{Conclusion}
In this paper a novel algorithm that reduces the computational complexity of the straightforward search in the DP approach is presented. The proposed method has been utilized for stereo and motion problems and has shown a significant speedup. Also a new expansion of the DP technique was proposed. This algorithm showed state-of-the-art results in terms of energy minimization and can be applied directly for many other early vision problems that use the MRF approach and the regular rectangular image grid.

\ifCLASSOPTIONcompsoc
 \section*{Acknowledgments}
\else
 \section*{Acknowledgment}
\fi

The author would like to thank the support of the Ramon y Cajal research program, MEC, Spain and the support of the EC grant IST-027110 for the HERMES 
project.

\ifCLASSOPTIONcaptionsoff
 \newpage
\fi


%





\end{document}